\documentclass[twocolumn,twoside]{IEEEtran}    

\usepackage{amsfonts}
\usepackage{amssymb}
\usepackage{amsbsy} 
\usepackage{amsmath}
\usepackage{cite}
\usepackage{subfigure}
\usepackage[dvips]{graphicx}
\usepackage[T1]{fontenc}
\usepackage[latin1]{inputenc}
\usepackage{hyperref}

\newcommand{\Actual}{\mathbb R}
\usepackage{float}

\def\x{{\mathbf x}}
\def\xk{{\mathbf x}^{(k)}}

\def\Pk{\boldsymbol{\Psi}_{(k)}}

\def\Rk{{\mathbf R}_{(k)}}
\def\I{{\mathbf I}}

\def\y{{\mathbf y}}

\newtheorem{theorem}{Theorem}[section]

\newtheorem{property}[theorem]{Property}
\newenvironment{proof}[1][Proof]{\begin{trivlist}
\item[\hskip \labelsep {\bfseries #1}]}{\end{trivlist}}

\usepackage{color}

\renewcommand{\baselinestretch}{0.98}

\begin{document}

\title{Iterative Gaussianization: \\ from {ICA} to Random Rotations}

\author{Valero Laparra, Gustavo Camps-Valls,~\IEEEmembership{Senior Member,~IEEE} and Jes\'us Malo
\thanks{Copyright (c) 2011 IEEE. Personal use of this material is permitted. Permission from IEEE must be obtained for all other users, including reprinting/
republishing this material for advertising or promotional purposes, creating new collective works for resale or redistribution to servers or
lists, or reuse of any copyrighted components of this work in other works. DOI: 10.1109/TNN.2011.2106511.}
\thanks{This work was partially supported by projects CICYT-FEDER
TEC2009-13696, AYA2008-05965-C04-03 and CSD2007-00018, and grant BES-2007-16125.}
\thanks{Authors are with Image Processing Laboratory (IPL) at the
Universitat de Val{\`{e}}ncia, Spain. Address:
C/ Catedr\'atico Escardino, 46980 - Paterna (Val{\`{e}}ncia) Spain. E-mail:
\{lapeva,gcamps,jmalo\}@uv.es, http://ipl.uv.es}}

\markboth{2011 \copyright IEEE Transactions on Neural Networks}{Laparra et al.: Iterative Gaussianization: from {ICA} to Random Rotations}

\maketitle

\begin{abstract}
Most signal processing problems involve the challenging task of
multidimensional probability density function (PDF) estimation.
In this work, we propose a solution to this problem by using a family of Rotation-based
Iterative Gaussianization (RBIG) transforms.
The general framework consists of the sequential application of a
univariate marginal Gaussianization transform followed
by an orthonormal transform.
The proposed procedure looks for differentiable transforms to a known PDF so
that the unknown PDF can be estimated at any point of the original domain.
In particular, we aim at a zero mean unit covariance Gaussian for convenience.

RBIG is formally similar to classical iterative Projection Pursuit (PP) algorithms.
However, we show that, unlike in PP methods, the particular
class of rotations used has no special qualitative relevance in this context, since looking
for {\em interestingness} is not a critical issue for PDF estimation.
The key difference is that our approach focuses on the univariate part (marginal Gaussianization) of the problem rather than on the multivariate part (rotation).
This difference implies that one may select the most
convenient rotation suited to each practical application.

The differentiability, invertibility and convergence of RBIG are theoretically
and experimentally analyzed. Relation to other methods, such as Radial Gaussianization (RG), one-class support
vector domain description (SVDD), and deep neural networks (DNN) is also pointed out.
The practical performance of RBIG is successfully illustrated in a number of multidimensional
problems such as image synthesis, classification, denoising, and multi-information
estimation.
\end{abstract}



\section{Introduction} \label{sec:introduction}

\IEEEPARstart{M}{any} signal processing problems such as coding, restoration, classification,
regression or synthesis greatly depend on an appropriate description of the underlying probability
density function (PDF) \cite{Gersho92,Banham97,Duda00,Hastie01,Portilla00}. However, density estimation is a
challenging problem when dealing with high-dimensional signals because direct sampling of the
input space is not an easy task due to the curse of dimensionality \cite{Scott92}. 
As a result, specific problem-oriented PDF models are typically developed to be used in the Bayesian framework.

The conventional approach is to transform data into a domain where \emph{interesting} features
can be easily (i.e. marginally) characterized.
In that case, one can apply well-known marginal techniques to each feature independently and
then obtain a description of the multidimensional PDF.
The most popular approaches rely on linear models and statistical independence. However, they are usually too
restrictive to describe general data distributions. For instance, principal component analysis (PCA) \cite{Jolliffe86},
that reduces to DCT in many natural signals such as speech, images and video, assumes a Gaussian
source \cite{Duda00,Jolliffe86}. More recently, linear ICA, that reduces to wavelets in natural signals,
assumes that observations come from the linear combination of independent non-Gaussian
sources \cite{Hyvarinen99b}. In general, these assumptions may not be completely correct,
and residual dependencies still remain after the linear transform that looks for independence.
As a result, a number of problem-oriented approaches have been developed in the last decade to
either describe or remove the relations remaining in these linear domains. For example, parametric models based
on joint statistics of wavelet coefficients have been successfully proposed for
texture analysis and synthesis \cite{Portilla00}, image coding \cite{Buccigrossi99} or image
denoising \cite{Portilla03}.
Non-linear methods using non-explicit statistical models have been also proposed to this end in
the denoising context \cite{Gutierrez06,Laparra10a} and in the coding context \cite{Malo06a,Camps08a}.
In function approximation and classification problems, a common approach is to first linearly
transform the data, e.g. with the most relevant eigenvectors from PCA, and then applying
nonlinear methods such as artificial neural networks or support vector machines in the reduced
dimensionality space \cite{Jolliffe86,Duda00,Hastie01}.

Identifying the \emph{meaningful} transform for an easier PDF description
in the transformed domain strongly depends on the problem at hand.
In this work we circumvent this constraint by looking for a transform such that the transformed PDF
is known. Even in the case that this transform is qualitatively \emph{meaningless}, being differentiable,
allows us to estimate the PDF in the original domain.
Accordingly, in the proposed context, the role (\emph{meaningfulness}) of the transform is not that relevant.
Actually, as we will see, an infinite family of transforms may be suitable to this end, so one has
the freedom to choose the most convenient one.

In this work, we propose to use a unit covariance Gaussian as target PDF in the transformed domain
and iterative transforms based on arbitrary rotations. We do so because the match between spherical symmetry
and rotations makes it possible to define a cost function (negentropy) with nice theoretical properties.
The properties of negentropy allow us to show that one Gaussianization transform is always found
no matter the selected class of rotations.

The remainder of the paper is organized as follows. In Section \ref{Motivation} we present the
underlying idea that motivates the proposed approach to Gaussianization.
In Section \ref{OurApproach}, we give the formal definition of the
Rotation-based Iterative Gaussianization (RBIG), and show that the scheme is invertible, differentiable and it converges
for a wide class of orthonormal transforms, even including random rotations.
Section \ref{RelationsPP} discusses the similarities and differences of the proposed
method and Projection Pursuit (PP) \cite{Friedman74,Huber85,Chen00,Rodriguez10}. Links to other
techniques (such as single-step Gaussianization transforms \cite{Erdogmus06,Lyu08c}, one-class
support vector domain descriptions (SVDD) \cite{Tax99}, and deep neural network architectures \cite{Hinton06})
are also explored.
Section \ref{applications} shows the experimental results.
First, we experimentally show
that the proposed scheme converges to an appropriate Gaussianization transform for a wide class of
rotations.
Then, we illustrate the usefulness of the method in a number of high-dimensional
problems involving PDF estimation: image synthesis, classification, denoising and multi-information estimation.
In all cases, RBIG is compared to related methods in each particular application.
Finally, Section \ref{Conclusions} draws the conclusions of the work.

\section{Motivation}
\label{Motivation}

{This section considers a solution to the PDF estimation problem
by using a differentiable transform to a domain with known PDF.
In this setting, different approaches can be adopted which will
motivate the proposed method.}

Let $\x$ be a $d$-dimensional random variable with (unknown) PDF, $p_{\mathbf{x}}(\mathbf{x})$.
Given some bijective, differentiable transform of $\x$ into $\y$, $\mathcal{G}:\Actual^d \to \Actual^d$,
such that $\mathbf{y}=\mathcal{G}(\mathbf{x})$, the PDFs in the original and the transformed domains
are related by \cite{Stark86}:

\begin{equation}
     p_{\x}(\x) = p_{\y}(\mathcal{G}(\x)) \bigg|\dfrac{d\mathcal{G}(\x)}{d\x}\bigg| = p_{\y}(\mathcal{G}(\x)) | \nabla_{\x} \mathcal{G}(\x) |,
     \label{stark}
\end{equation}
where $|\nabla_{\mathbf{x}} \mathcal{G}|$ is the determinant of the Jacobian matrix.
Therefore, the unknown PDF in the original domain can be estimated from a transform of known Jacobian
leading to an appropriate (known or straightforward to compute) target PDF, $p_{\y}(\y)$.

One could certainly try to figure out direct (or even closed form) procedures to transform particular
PDF classes into a target PDF \cite{Erdogmus06,Lyu08c}. However, in order to deal with any possible PDF, iterative
methods seem to be a more reasonable approach. In this case, the initial data distribution should be iteratively
transformed in such a way that the target PDF is progressively approached in each iteration.

The appropriate transform in each iteration would be the one that maximizes a similarity measure between
PDFs. A sensible cost function here is the Kullback-Leibler divergence (KLD) between PDFs.
In order to apply well-known properties of this measure \cite{Comon94,Cardoso03}, it is convenient to choose
a unit covariance Gaussian as target PDF, $p_{\y}(\y)=\mathcal{N}(\mathbf{0},\mathbf{I})$.
With this choice, the cost function describing the divergence between the current data, $\x$, and the unit covariance
Gaussian is the hereafter called negentropy\footnote{This usage of the term negentropy slightly differs from the
usual definition \cite{Comon94} where negentropy is taken to be KLD between $p_{\x}(\x)$ and a multivariate
Gaussian of the same mean and covariance. However, note that this difference has no consequence assuming the
appropriate input data standardization (zero mean and unit covariance), which can be done without loss of
generality.}, $J(\x)$ = $\text{D}_{\text{KL}}\left(p(\x) | \mathcal{N}({\bf 0},\I)\right)$.
Negentropy can be decomposed as the sum of two non-negative quantities, the multi-information and the
marginal negentropy:
\begin{equation}
     J(\mathbf{x}) = I(\mathbf{x}) + J_m(\mathbf{x}).
\label{negentropy-descomposition1}
\end{equation}
This can be readily derived from Eq. (5) in \cite{Cardoso03}, by considering as contrast PDF
$\prod_{i} q_i(x_i)=\mathcal{N}(\mathbf{0},\mathbf{I})$.
The multi-information is \cite{Studeny98}:
\begin{equation} \textstyle
    I(\x) = \text{D}_{\text{KL}}( p(\x) | \prod_{i} p_i(x_i))
    \label{multiinformation}
\end{equation}
Multi-information measures statistical dependence, and
it is zero if and only if the different components of $\x$ are independent.
The marginal negentropy is defined as:
\begin{equation}
    J_m(\x) = \sum_{i=1}^{d} \text{D}_{\text{KL}}\left(p_i(x_i) | \mathcal{N}(0,1)\right)
    \label{marginalnegentropy}
\end{equation}

Given a data distribution from the unknown PDF, in general both $I$ and $J_m$ will be non-zero.
The decomposition in \eqref{negentropy-descomposition1} suggests two alternative approaches to reduce $J$:
\begin{enumerate}
  \item \emph{Reducing $I$}: This implies looking for interesting (independent) components.
        If one is able to obtain $I=0$, then $J = J_m \geq 0$, and this reduces to
        solving a marginal problem. Marginal negentropy can be set to zero with the appropriate set
        of dimension-wise Gaussianization transforms, $\mathbf{\Psi}$. This is easy as will be shown in the next section.

        However, this is an ambitious approach since looking for independent components is a non-trivial (intrinsically multivariate and nonlinear) problem. According to this, linear ICA techniques will not succeed in completely removing the
        multi-information, and thus a nonlinear post-processing is required.
  \item \emph{Reducing $J_m$}: As stated above, this univariate problem is easy to solve by using the appropriate $\mathbf{\Psi}$.
        Note that $I$ will remain constant since it is invariant under dimension-wise transforms \cite{Studeny98}.
        In this way, one ensures that the cost function is reduced by $J_m$.
        Then, a further processing has to be taken in order to come back to a situation in which one
        may have the opportunity to remove $J_m$ again. This additional transform may consist of applying a
        rotation $\mathbf{R}$ to the data, as will be shown in the next section.
\end{enumerate}
The relevant difference between the approaches is that, in the first one, the important part is looking for the interesting representation (multivariate problem), while in the second approach the important part is the univariate Gaussianization. In this second case, the class of rotations has no special qualitative relevance: in fact,
marginal Gaussianization is the only part reducing the cost function.

\begin{figure*}[t!]
 \vspace{-0.1cm}
 \hspace{-0.6cm}
 \begin{tabular}{ccccc}
 \includegraphics[scale=0.43]{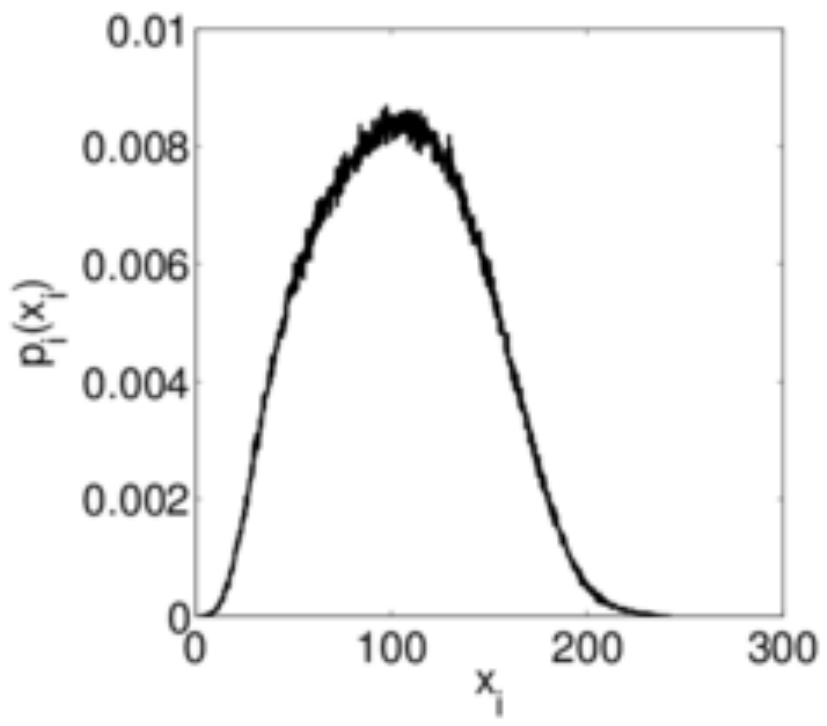} \hspace{-0.3cm} & \hspace{-0.3cm}
 \includegraphics[scale=0.43]{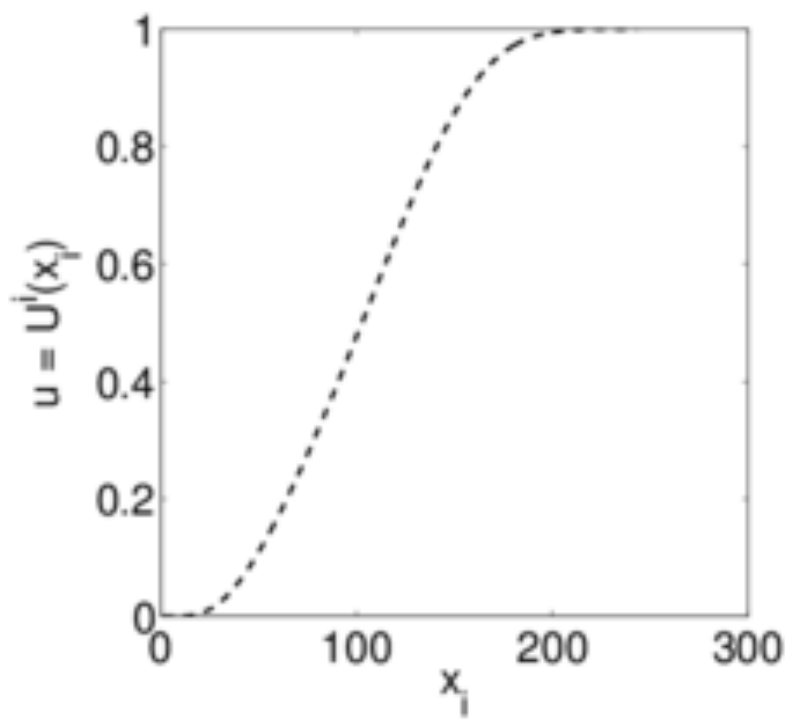} \hspace{-0.3cm} & \hspace{-0.3cm}
 \includegraphics[scale=0.43]{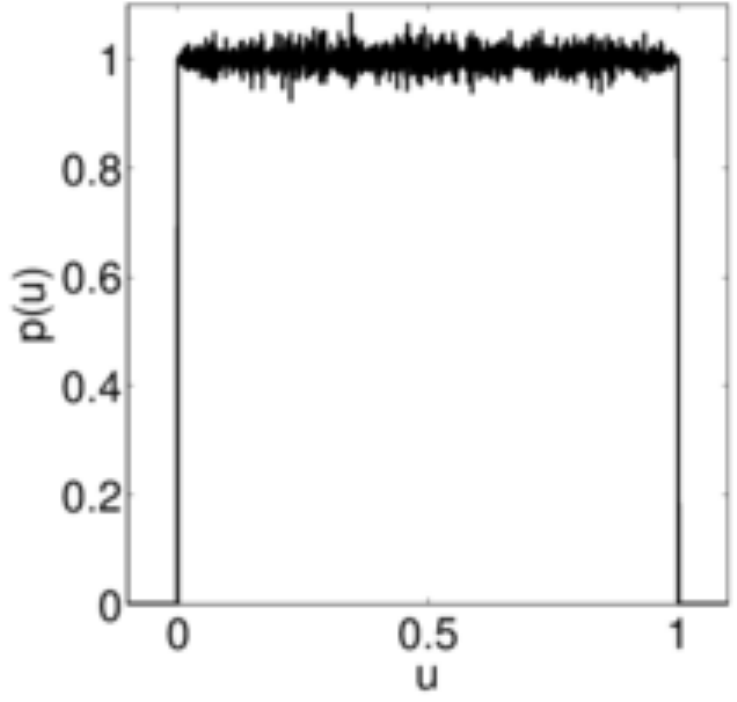} \hspace{-0.3cm} & \hspace{-0.3cm}
 \includegraphics[scale=0.43]{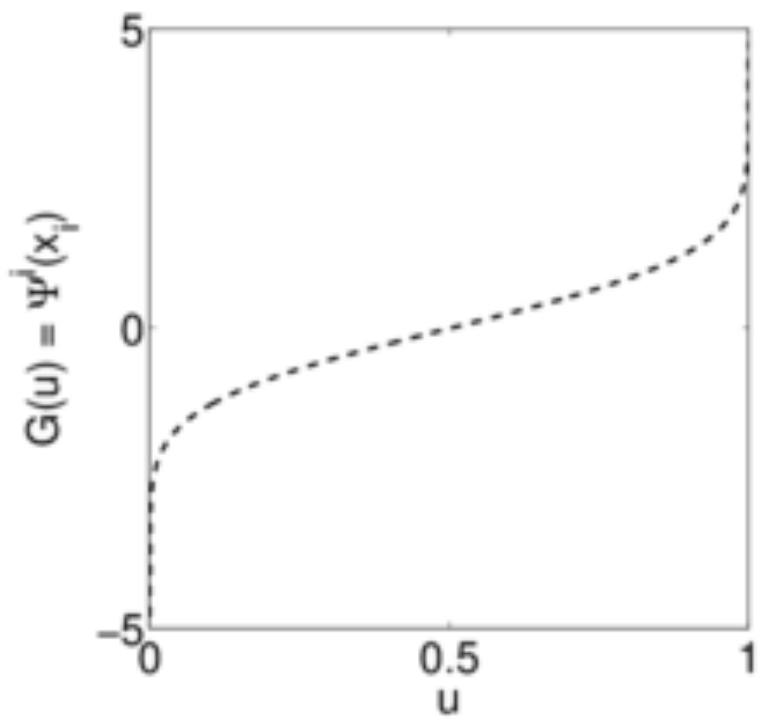} \hspace{-0.3cm} & \hspace{-0.3cm}
 \includegraphics[scale=0.43]{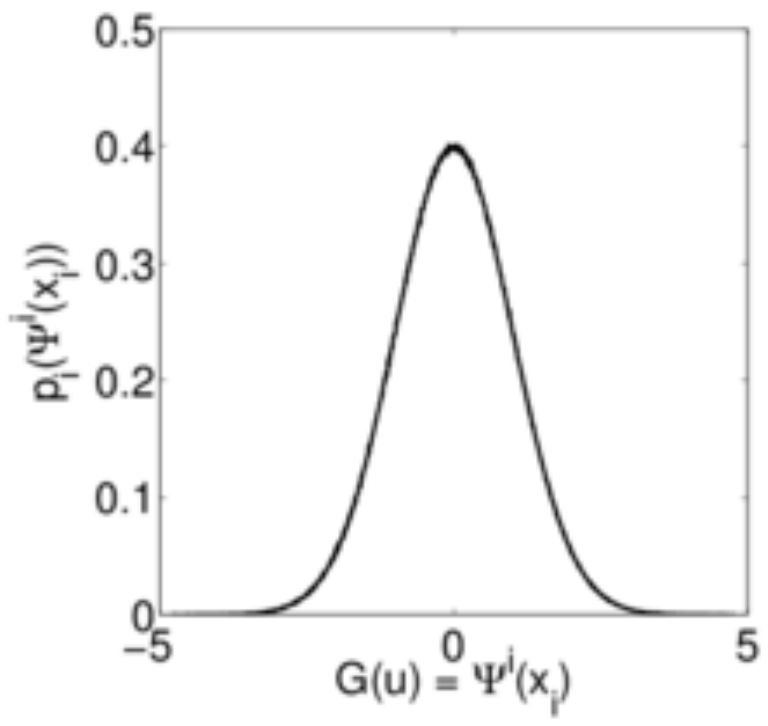}
\end{tabular}
 \caption{Example of marginal Gaussianization in some particular dimension $i$. From left to right: marginal PDF of $x_i$, uniformization transform $u=U^i(x_i)$, PDF of the uniformized variable $p(u)$, Gaussianization transform $G(u)$, and PDF of the Gaussianized variable $p_i(\Psi^i(x_i))$.}
 \label{example_marg_gauss}
\end{figure*}

The first approach is the underlying idea in Projection Pursuit methods focused on
looking for interesting projections \cite{Huber85,Chen00}.
Since the core of these methods is looking for meaningful projections (usually ICA algorithms),
they suffer from a big computational complexity: for example, robust ICA algorithms
such as RADICAL \cite{LearnedMiller03} would lead to extremely slow Guassanization algorithms
whereas relatively more convenient alternatives such as FastICA \cite{Hyvarinen99}
may not converge in all cases.
This may explain why, so far, Gaussianization techniques have been applied just to
low-dimensional (audio) signals in either simple contexts based on point-wise nonlinearities \cite{Zhang05,Squartini06},
or after {\em ad hoc} speech-oriented feature extraction steps \cite{Xiang02}.
In this work, we propose following
the simpler second approach using the most computationally
convenient rotation. Intentionally, we do not pay attention to the meaningfulness of
the rotations.

\section{Rotation-based Iterative Gaussianization (RBIG)}
\label{OurApproach}

This section first introduces the basic formulation of the proposed method, and then
analyzes the properties of differentiability, invertibility, and convergence. Finally,
we discuss on the role of the rotation matrix used in the scheme.

\subsection{Iterative Gaussianization based on arbitrary rotations}

According to the above reasoning, we propose the following class of Rotation-based Iterative Guassianization (RBIG) algorithms:
given a $d$-dimensional random variable $\x^{(0)}$, following an unknown PDF,
$p(\x^{(0)})$, in each iteration $k$, a two-step processing is performed:
\begin{equation}\label{gauss}
 {\mathcal G}: \x^{(k+1)} = \Rk \cdot \Pk(\xk)
\end{equation}
where $\Pk$ is the marginal Gaussianization of each dimension of $\xk$ for the corresponding iteration,
and $\Rk$ is a generic rotation matrix for the marginally Gaussianized variable $\Pk(\xk)$.

The freedom in choosing the rotations is consistent with the intuitive fact that there
is an infinite number of ways to twist a PDF in order to turn it into a unit covariance Gaussian.
In principle, any of these choices is equally useful for our purpose, i.e. estimating the PDF in
the original domain using Eq. \eqref{stark}.
Note that when using different rotations,
the qualitative meaning of the same region of the corresponding Gaussianized domain will be different.
As a result, in order to work in the Gaussianized domain, one has to take into
account the value of the point-dependent Jacobian.
Incidentally, this is also the case in the PP approach, and more generally, in any
non-linear approach.
However, the interpretation of the Gaussianized domain is not an issue when working in the
original domain. Finally, it is important to note that the method just depends on univariate
(marginal) PDF estimations. Therefore, it does not suffer from the curse of dimensionality.

\subsection{Invertibility and differentiation}

The considered class of Gaussianization transforms is {\em differentiable} and {\em invertible}.
Differentiability, allows us to estimate the PDF in the original domain from the
Jacobian of the transform in each point, cf. Eq. \eqref{stark}.
Invertibility guarantees that the transform is bijective which is a necessary condition to apply
Eq. \eqref{stark}. Additionally, it is convenient for generating samples in the original domain by sampling
the Gaussianized domain.

Before getting into the details, we take a closer look at the basic tool of marginal Gaussianization.
Marginal Gaussianization in each dimension $i$ and each iteration $k$, $\Pk^i$, can be decomposed into two equalization transforms: (1) marginal uniformization, $U^i_{(k)}$, based on the cumulative density function of the marginal PDF, and (2) Gaussianization of a uniform variable, $G(u)$, based on the inverse of the cumulative density function of a univariate Gaussian: $\Pk^{i} = G \odot U^i_{(k)}$, where:
\begin{eqnarray}\label{lau}
      u = U^i_{(k)}(x_i^{(k)})&=&\int_{-\infty}^{x_i^{(k)}} p_i(x_i^{\prime (k)}) \, dx_i^{\prime (k)} \\
      G^{-1}(x_i) &=&\int_{-\infty}^{x_i} g(x^{\prime}_i) \, dx^{\prime}_i
      \label{marginal_gauss}
\end{eqnarray}
and $g(x_i)$ is just a univariate Gaussian.
Figure \ref{example_marg_gauss} shows an example of the marginal Gaussianization of a one-dimensional variable $x_i$.

One dimensional density estimation is an issue by itself, and it has been widely studied \cite{Silverman86, Hastie01}. The selection of the most convenient density estimation procedure depends on the particular problem and, of course, the univariate Gaussianization step in the proposed algorithm could benefit from the extensive literature on the issue. In our case, we take a practical approach and no particular model is assumed for the marginal variables to keep the method as general as possible. Accordingly, the univariate Gaussianization transforms are computed from the cumulative histograms. Of course, alternative analytical approximations could be introduced at the cost of making the model more rigid. On the positive side, parametric models may imply better data regularization and avoid overfitting. However, exploring the effect of alternative density estimators will not be analyzed here.

Let us consider now the issue of invertibility.
By simple manipulation of \eqref{gauss}, it can be shown that the inverse transform is given by:
\begin{equation}\label{invgauss}
 {\mathcal G}^{-1}: \x^{(k)} = \Pk^{-1}(\Rk^\top \cdot \x_{(k+1)}).
\end{equation}
The rotation $\Rk$ is not a problem for invertibility since the inverse is just the transpose, $\Rk^{-1}=\Rk^\top$.
However, the key to ensure transform inversion is the invertibility of $\Pk$. This is trivially ensured when the support of each marginal PDF is connected, that is, there are no holes (zero probability regions) in the support. In this way all the marginal CDFs are strictly monotonic and hence invertible. Note that the existence of holes in the support of the joint PDF is not a problem as long as it gives rise to marginal PDFs with a
connected support.
Problems in inversion will appear only when the joint PDF gives rise to clusters that are so distant that their projections onto the axes do not overlap. However, in such a situation, it may make more qualitative sense to consider that distinct clusters come from different sources and learn each one with a different Gaussianization transform.

The Jacobian of the series of $K$ iterations is just the product of the corresponding Jacobian in each iteration:
\begin{equation}\textstyle
      \nabla_{\mathbf{x}} \mathcal{G} = \prod_{k=1}^{K} \Rk  \cdot \nabla_{\xk} \Pk
\end{equation}
Marginal Gaussianization, $\Pk$, is a dimension-wise transform, whose Jacobian is the diagonal matrix,
\begin{equation}
      \nabla_{\xk} \Pk = \begin{pmatrix}
                                  \dfrac{\partial \Pk^1}{\partial x_1^{(k)}} & \cdots & 0 \\
                                  \vdots & \ddots & \vdots \\
                                  0 & \cdots & \dfrac{\partial \Pk^d}{\partial x_d^{(k)}} \\
                                \end{pmatrix}
\end{equation}
According to the two equalization steps in each marginal Gaussianization, Eq. \eqref{marginal_gauss}, each element in $\nabla_{\xk} \Pk$ can be easily computed by applying the chain rule on $u$ defined in Eq. \eqref{lau}:
\begin{eqnarray}\nonumber
      \dfrac{\partial \Pk^i}{\partial x_i^{(k)}} & = & \dfrac{\partial {\mathcal G}}{\partial u}  \dfrac{\partial u }{\partial x_i^{(k)}} = \left(\dfrac{\partial {\mathcal G}^{-1}}{\partial x_i} \right)^{-1}  p_i(x_i^{(k)}) \\
& = & g(\Psi^i_{(k)}(x_i^{(k)}))^{-1}  p_i(x_i^{(k)})
      \label{elem_jacobiano}
\end{eqnarray}
Again, the differentiable nature of the considered Gaussianization is independent from the selected rotations $\Rk$.

\subsection{Convergence properties}
\label{Convergence properties}

Here we prove two general properties of random variables, which are useful in the contexts of PDF
description and redundancy reduction.

\begin{property}[Negentropy reduction]\label{property1}
Marginal Gaussianization reduces the negentropy and this is not modified by any posterior rotation:
\begin{equation}
    \Delta J = J(\mathbf{x}) - J(\mathbf{R} \mathbf{\Psi}(\mathbf{x})) \geq 0, \, \forall \, \mathbf{R}
    \label{prop1}
\end{equation}
\end{property}

\begin{proof}
Using Eq. \eqref{negentropy-descomposition1}, the negentropy reduction due
to marginal Gaussianization followed by a rotation is:
\begin{equation} \nonumber
      \Delta J = J(\mathbf{x}) - J(\mathbf{R} \mathbf{\Psi}(\mathbf{x})) = J(\mathbf{x}) - J(\mathbf{\Psi}(\mathbf{x}))
\end{equation}
since $\mathcal{N}({\bf 0,I})$ is rotation invariant. Therefore,
\begin{eqnarray} \nonumber
  \Delta J & = & I(\mathbf{x}) + J_m(\mathbf{\mathbf{x}}) - I(\mathbf{\Psi}(\mathbf{x})) - J_m(\mathbf{\Psi}(\mathbf{x}))
\end{eqnarray}
Since the multi-information is invariant under dimension-wise transforms \cite{Studeny98} (such as $\mathbf{\Psi}$), and the marginal negentropy of a marginally Gaussianized variable is zero,
\begin{eqnarray} \nonumber
     \Delta J = J_m( \mathbf{x} ) \geq 0, \, \forall \, \mathbf{R}
\label{prop1_explicit}
\end{eqnarray}
\vspace{-0.07cm}
\end{proof}

\begin{property}[Redundancy reduction]\label{property2}
Given a marginally Gaussianized variable, $\mathbf{\Psi}(\mathbf{x})$, any rotation reduces the redundancy among coefficients,
\begin{equation}
      \Delta I = I(\mathbf{\Psi}(\mathbf{x})) - I(\mathbf{R} \mathbf{\Psi}(\mathbf{x})) \geq 0, \, \forall \, \mathbf{R}
      \label{prop2}
\end{equation}
Note that this property also implies that the combination of marginal Gaussianization and rotation gives rise to redundancy
reduction since $I(\mathbf{\Psi}(\mathbf{x}))=I(\mathbf{x})$.
\end{property}

\begin{proof}
Using Eq. \eqref{negentropy-descomposition1} on both $I(\mathbf{\Psi}(\mathbf{x}))$ and $I(\mathbf{R} \mathbf{\Psi}(\mathbf{x}))$, the redundancy reduction is:
\begin{eqnarray} \nonumber
\Delta I = J(\mathbf{\Psi}(\mathbf{x})) - J_m(\mathbf{\Psi}(\mathbf{x})) - J(\mathbf{R} \mathbf{\Psi}(\mathbf{x})) + J_m(\mathbf{R} \mathbf{\Psi}(\mathbf{x})). \nonumber
\end{eqnarray}
Since negentropy is rotation invariant and the marginal negentropy of a marginally Gaussianized variable is zero,
\begin{eqnarray} \nonumber
      \Delta I = J_m(\mathbf{R} \mathbf{\Psi}(\mathbf{x})) \geq 0, \, \forall \, \mathbf{R}
      \label{prop2_explicit}
\end{eqnarray}

\vspace{-0.07cm}
\end{proof}
The above properties suggest the convergence of the proposed Gaussianization method.
Property \ref{property1} (Eq. \eqref{prop1}) ensures that the distance between
the PDF of the transformed variable to a zero mean unit covariance multivariate Gaussian is
reduced in each iteration. Property \ref{property2} (Eq. \eqref{prop2})
ensures that redundancy among coefficients is also reduced after each iteration.
According to this the distance to a Gaussian will decay to zero for a wide class of rotations.

\subsection{On the rotation matrices}
\label{On the rotation matrices}

Admissible rotations are those that change the situation after marginal Gaussianization in
such a way that $J_m$ is increased.
Using different rotation matrices gives rise to different properties of the algorithm.

The above Properties \ref{property1} and \ref{property2} provide some intuition on the suitable
class of rotations. By using \eqref{prop1} and \eqref{prop2} in the sequence \eqref{gauss}, one readily obtains the relations:
\begin{equation} \label{chocante}
\Delta J_{(k)} = J_m(\xk) = \Delta I_{(k-1)},
\end{equation}
and thus, interestingly, the amount of negentropy reduction (the convergence rate) at some iteration $k$ will
be determined by the amount of redundancy reduction obtained in the previous iteration, $k-1$.
Since dependence can be analyzed
in terms of correlation and non-Gaussianity \cite{Cardoso03}, the intuitive candidates for $\mathbf{R}$
include orthonormal ICA, hereafter simply referred to as ICA, which maximizes the redundancy reduction;
and PCA, which removes correlation.
Random rotations (RND) will be considered here as an extreme case to point out
that looking for interesting projections is not critical to achieve convergence.
Note that other rotations are possible, for instance, a quite sensible choice would be
randomly selecting projections that uniformly recover the surface of an hypersphere \cite{Leon06}.
Other possibilities include extension to complex variables \cite{Novey08}.

As an illustration, Table \ref{tablon} summarizes the main characteristics of the method when
using ICA, PCA and RND.
The table analyzes the closed-form nature of each rotation, the theoretical convergence of the
method, the convergence rate (negentropy reduction {\em per} iteration), and the computational cost
of each rotation. Section \ref{Convergenceexample} is devoted to the experimental confirmation of the
reported characteristics of convergence presented here.

\begin{table}[b!]
\footnotesize
\caption{Properties of the Gaussianization method for different rotations (see comments in the text).}
\vspace{-0.5cm}
\begin{center}
\setlength{\tabcolsep}{4pt}
\begin{tabular}{c||c|c|c|c}
\hline
               & Closed & Theoretical & Convergence  & CPU cost$^\dag$\\
Rotation       &  -form & convergence & rate         & \cite{zarzoso06,Sharma07,Golub96}\\
\hline
ICA            &   $\times$  &       $\surd$        &  Max $\Delta J$          &   ${\mathcal O}(2md(d+1)n)$     \\
PCA            &   $\surd$   &       $\surd$       &  $\Delta J$ = 2nd order  &   ${\mathcal O}(d^2(d+1)n)$     \\
RND            &   $\surd$   &       $\surd$        &  $\Delta J \geq 0$       &   ${\mathcal O}(d^3)$     \\
\hline
\end{tabular}
\end{center}
{\footnotesize $^\dag$ Computational cost considers $n$ samples of dimension $d$. The cost for the
ICA transform is that of FastICA running $m$ iterations.}
\label{tablon}
\end{table}

Using ICA guarantees the theoretical convergence of the Gaussianization process
since it seeks for the maximally non-Gaussian marginal PDFs. Therefore, the negentropy reduction
$\Delta J$ (Eq. \eqref{prop1}) is always strictly positive,
except for the case that the Gaussian PDF has been finally achieved.
This is consistent with previously reported results \cite{Chen00}.
Moreover, the convergence rate is optimal for ICA since it gives rise to the maximum $J_m(\x)$ (indicated in Table \ref{tablon} with `Max $\Delta J$').
However, the main problem of using ICA as the rotation matrix is that it has no closed-form solution, so
ICA algorithms typically resort to iterative procedures with either difficulties
in convergence or high computational load.

Using PCA leads to sub-optimal convergence rate because
it removes second-order redundancy
(indicated in Table \ref{tablon} with `$\Delta J$ = 2nd order'),
but it does not
maximize the marginal non-Gaussianity $J_m(\x)$.
Using PCA guarantees the convergence for every input PDF except for one singular case:
consider a variable $\xk$ which is not Gaussian but all
its marginal PDFs are univariate Gaussian and with a unit covariance matrix. In this case, $\Delta J_{(k+1)}=J_m(\xk)=0$, i.e. no
approximation to the Gaussian in negentropy terms is obtained in the next iteration. Besides, since
$\mathbf{\Psi}_{(k+1)}(\xk)=\xk$, the next PCA, $\mathbf{R}_{(k+1)}$, will be the identity matrix, thus
$\mathbf{x}^{(k+1)}=\xk$: as a result, the algorithm may get stuck into a negentropy local minimum.
In our experience, this undesired effect never happened in real datasets.
On the other hand, advantages of using PCA is that the solution
is closed-form, very fast, and even though the convergence rate is lower than for ICA,
the solution is achieved in a fraction of the time.

Using RND transforms guarantees the theoretical convergence of the method since random rotations
ensure that, even in the above considered singular case, the algorithm will not
be stuck into this particular non-Gaussian solution. On the contrary, if the achieved marginal
non-Gaussianity is zero after an infinite number of random rotations, it is because the
desired Gaussian solution has been finally achieved (Cramer-Wold Theorem \cite{Feller68}).
In practice, the above property of RND can be used as a way to check convergence when
using other rotations (e.g. PCA): when the zero marginal non-Gaussianity situation is
achieved, a useful safety check consists of including RND-based iterations.
In the RND case, the convergence rate is clearly sub-optimal, yet non-negative ($\Delta J \geq 0$):
the amount of negentropy reduction may take any value between zero and the maximum achieved by ICA. However, the method is much
faster in practice: even though it may take more iterations to converge, the cost of each
transform does not depend on the number of samples. The rotation matrix can be computed by fast orthonormalization
techniques \cite{Golub96}.
In this case, the computation time of the rotation is negligible compared
to that of the marginal Gaussianization.


\section{Relation to other methods}
\label{RelationsPP}

In this section we discuss the relation of RBIG to previously reported Gaussianization methods, including
iterative PP-like techniques \cite{Friedman74,Huber85,Chen00} and direct approaches suited for particular PDFs \cite{Erdogmus06,Lyu08c,Eichhorn09}. Additionally, relations to other machine learning tools, such as Support Vector Domain Description \cite{Tax99} and deep neural networks \cite{Hinton06} are also considered.

\subsection{Iterative Projection Pursuit Gaussianization}

As stated above, the aim of Projection Pursuit (PP) techniques \cite{Friedman74,Huber85} is looking
for interesting linear projections according to some projection index
measuring interestingness, and \emph{after}, this
interestingness is captured by removing it through the appropriate marginal equalization, thus making a step
from structure to disorder.
When interestingness or structure is defined by departure from disorder, non-Gaussianity or negentropy,
PP naturally leads to iterative application of non-orthogonal ICA transforms followed by marginal
Gaussianization, as in \cite{Chen00}:
\begin{equation}\label{order_to_disorder}
 {\mathcal G}: \x^{(k+1)} = \Pk( \mathbf{R}^{\mathrm{ICA}} \cdot \xk)
\end{equation}
As stated in Section \ref{Motivation}, this is \emph{Approach 1} to the Gaussian goal.
Unlike PP, RBIG aims at the Gaussian goal following \emph{Approach 2}.
The differences between \eqref{order_to_disorder} and \eqref{gauss} (reverse order between
the multivariate and the univariate transforms) suggest the different qualitative weight given
to each counterpart. While PP gives rise to an \emph{ordered} transition from structure to
disorder\footnote{In PP the structure of the unknown PDF in the input domain is progressively removed in each iteration starting from the most relevant projection and continuing by the second one, and so on, until total disorder (Gaussianity) is achieved.}, RBIG follows a \emph{disordered} transition to disorder.

\subsection{Direct (single-iteration) Gaussianization algorithms}

Direct (non-iterative) Gaussianization approaches are possible if the method has to be applied to restricted classes of PDFs, for example: (1) PDFs that can be marginally Gaussianized in the \emph{appropriate axes} \cite{Erdogmus06}, or (2) elliptically
symmetric PDFs so that the final Gaussian can be achieved by equalizing
the length (norm) of the whitened samples \cite{Lyu08c,Eichhorn09}.

The method proposed in \cite{Erdogmus06} is useful when combined with tools that can identify marginally Gaussianizable components, somewhat related to ICA transforms.
Nevertheless, the use of alternative transformations is still an open issue.
Erdogmus et al. proposed PCA, vector quantization or clustering as alternatives to ICA in order to find the most potentially `Gaussianizable' components. In this sense, the method could be seen as a particular case of PP in that it only uses one iteration: first finding the most appropriate representation and then using marginal Gaussianization.
Elliptically symmetric PDFs constitute a relevant class of PDFs in image processing applications
since this kind of functions is an accurate model of natural images (e.g. Gaussian Scale Mixtures \cite{Portilla03} and related models \cite{Malo10} share this symmetry).
Radial Gaussianization (RG) was specifically developed to deal with these particular kind of models \cite{Lyu08c}.
This transform consists of
a nonlinear function that acts radially, equalizing the histogram of the magnitude (energy) of the data to obtain the histogram
of the magnitude of a Gaussian.
Other methods
have exploited this kind of transformation to generalize it to $L_p$ symmetric distributions \cite{Eichhorn09}.
Obviously, elliptical symmetry is a fair assumption for natural images, but it may not be appropriate for other problems. Even in the image context, particular images may not strictly follow distributions with elliptical symmetry, therefore if RG-like transforms are applied to these images, they will give rise to non-Gaussianized data.

\begin{figure}[t!]
\begin{center}
\setlength{\tabcolsep}{4pt}
 \begin{tabular}{cccc}
Image & Image PDF & RG & RBIG \\
\hline
 & (0.34) & (0.04) & (0.0006) \\
 \includegraphics[scale=0.24]{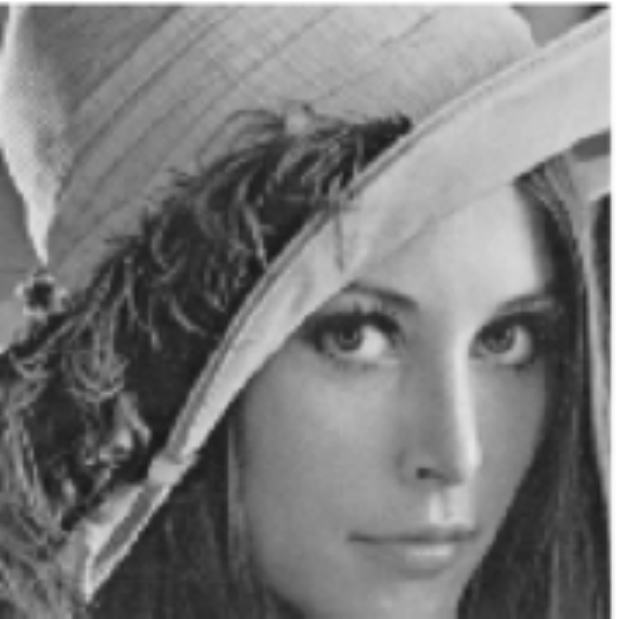} \hspace{-0.3cm} &
 \includegraphics[scale=0.24]{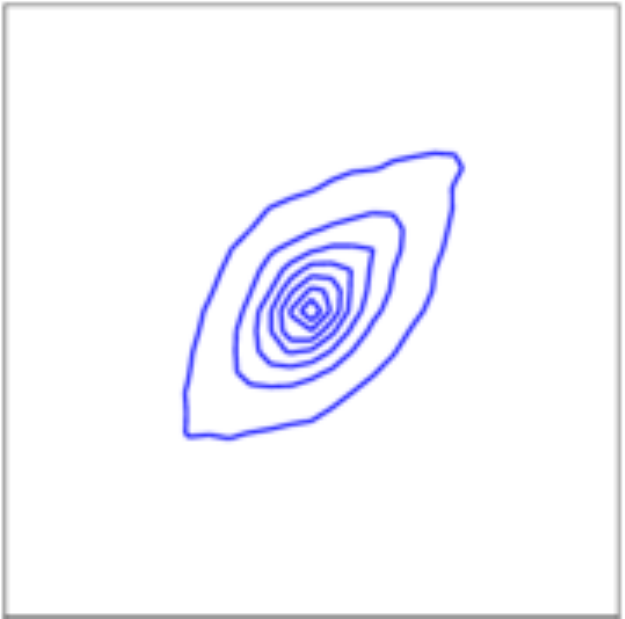} \hspace{-0.5cm} &
 \includegraphics[scale=0.24]{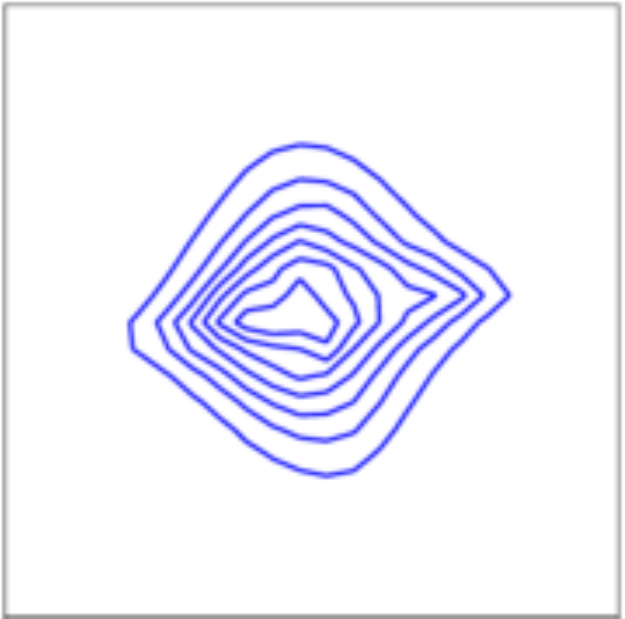} \hspace{-0.5cm} &
 \includegraphics[scale=0.24]{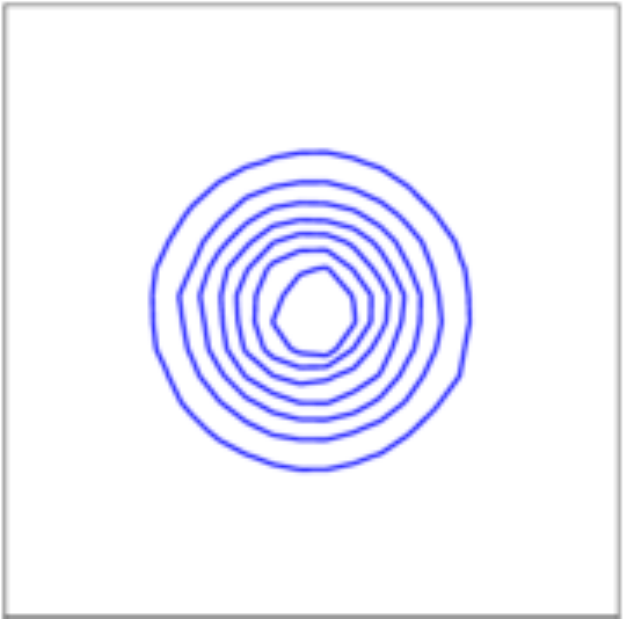}\\
& (0.59) & (0.034) & (0.0002) \\
 \includegraphics[scale=0.24]{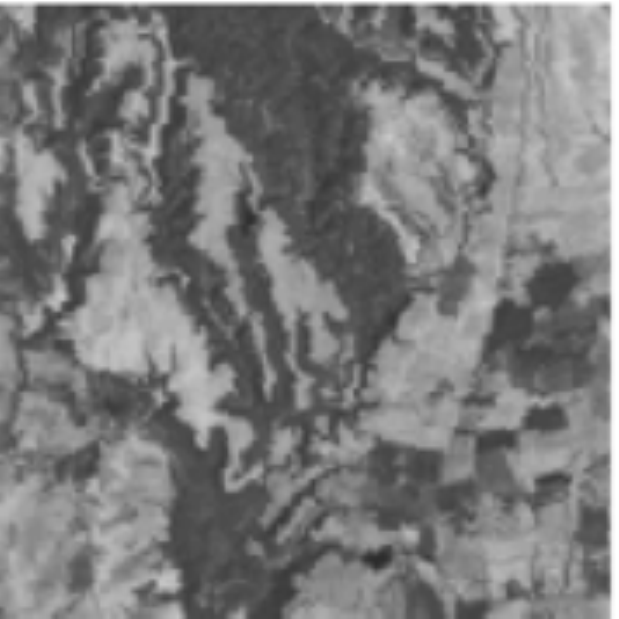} \hspace{-0.3cm} &
 \includegraphics[scale=0.24]{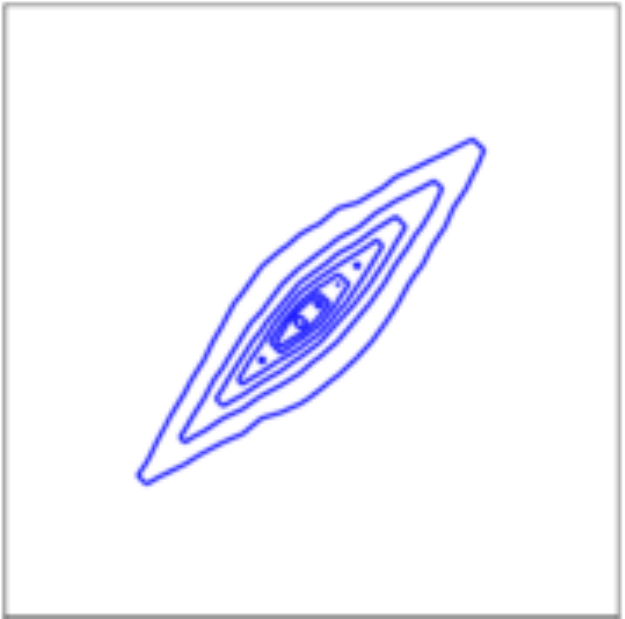} \hspace{-0.5cm} &
 \includegraphics[scale=0.24]{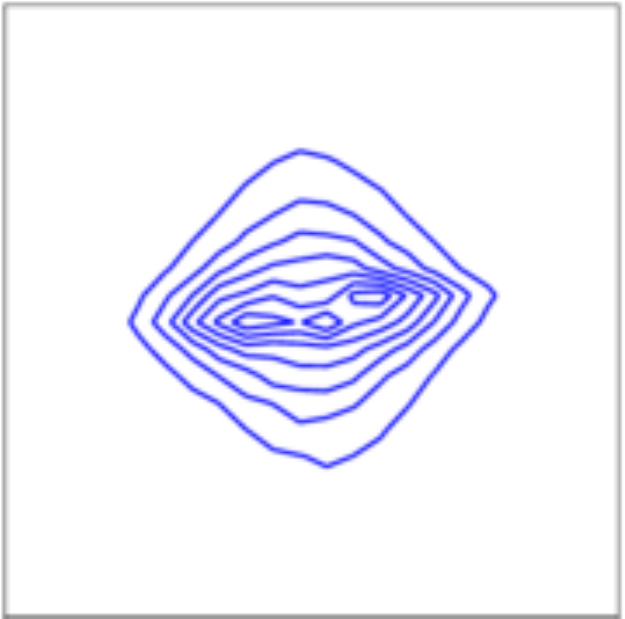} \hspace{-0.5cm} &
 \includegraphics[scale=0.24]{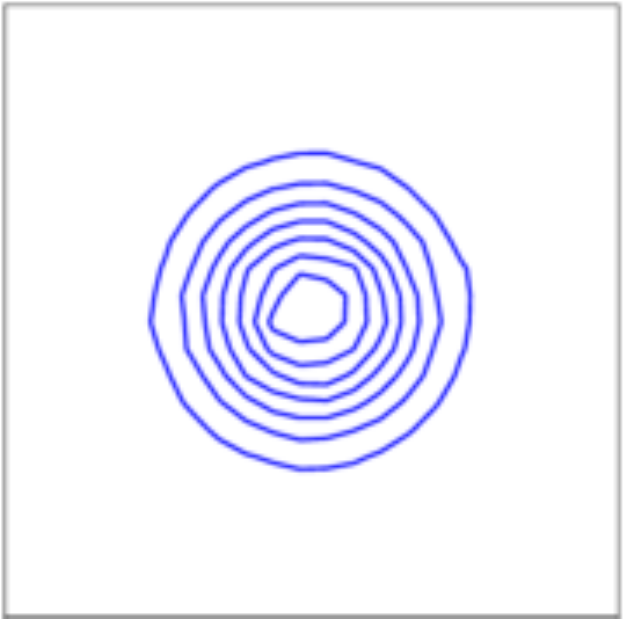}\\
&(0.066) & (0.052) & (0.0001) \\
 \includegraphics[scale=0.24]{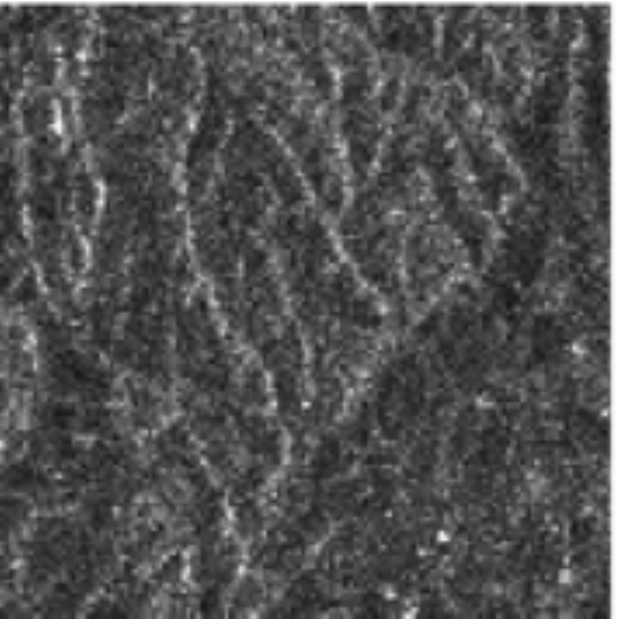} \hspace{-0.3cm} &
 \includegraphics[scale=0.24]{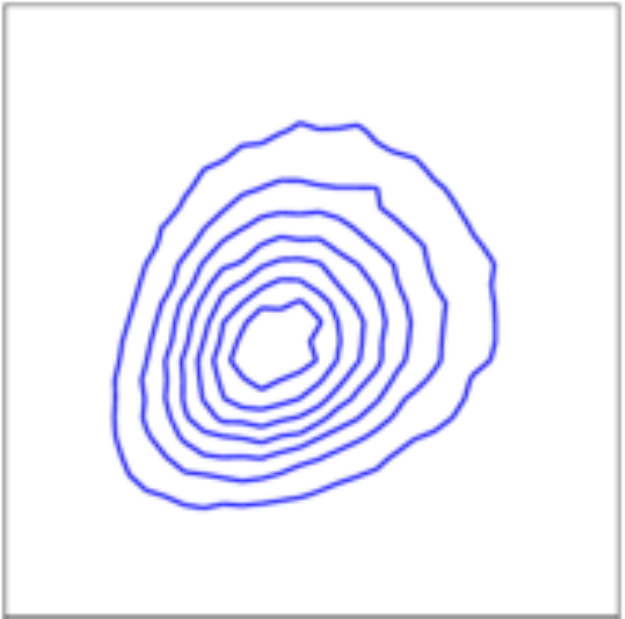} \hspace{-0.5cm} &
 \includegraphics[scale=0.24]{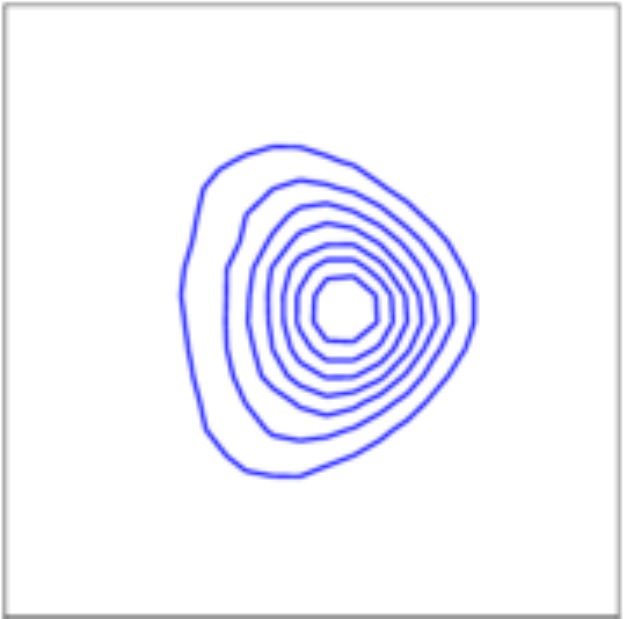} \hspace{-0.5cm} &
 \includegraphics[scale=0.24]{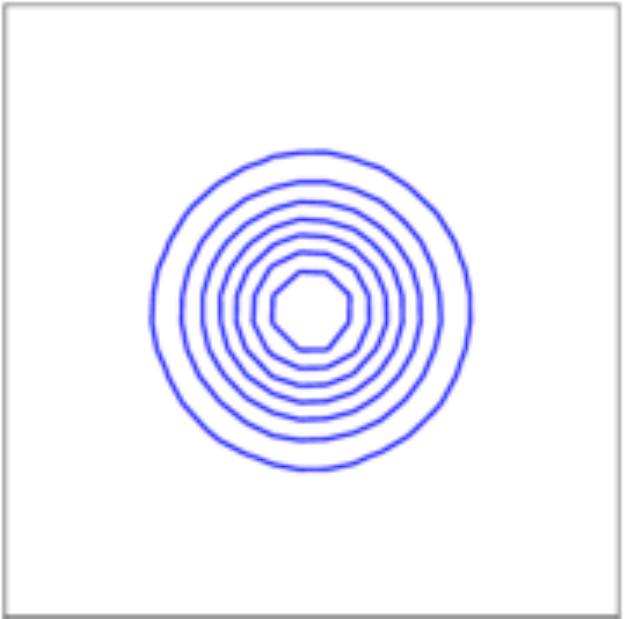}\\
 \end{tabular}
\end{center}
 \caption{Gaussianization of pairs of neighbor pixels from different images with RG and RBIG: natural image (top row), remote sensing Landsat channel in the optical range (middle row) and intensity of a ERS2 synthetic aperture radar (SAR) image (bottom row). Contour plots show the PDFs in the corresponding domains. The estimated mutual information (in bits) is given in parenthesis.}
\label{fig:Nat_Im_RG}
\end{figure}

Figure \ref{fig:Nat_Im_RG} shows
this effect in three types of acquired images: (1) a standard grayscale image, i.e. a typical example of a natural photographic image, (2) a band (in the visible range) of a remote sensing
multispectral image acquired by the Landsat sensor, and (3) a ERS2 synthetic aperture radar (SAR) intensity image for the same scene (of course out of the visible range).
In these illustrative examples, RG and RBIG were trained with the data distribution of pairs of neighbor pixels for each image, and RBIG was implemented using PCA rotations according to the results in Section \ref{Convergenceexample}.
Both RG and RBIG strongly reduce the mutual-information of pairs of neighbor pixels (see the mutual information values, in bits),
but it is noticeable that RG is more effective, higher $I$ reduction, in the natural image cases (photographic and visible channel images),
in which the assumption of elliptically symmetric PDF is more reliable.
However, it obviously fails when considering non-natural (radar) images, far from the visible range ($I$ is not significantly reduced).
The proposed method is more robust to these changes in the underlying PDF because no assumption is made.

\subsection{Relation to Support Vector Domain Description}\label{relsvdd}

The Support Vector Domain Description (SVDD) is a one-class classification method that finds a minimum volume sphere in a kernel feature space that contains $1-\nu$ fraction of the \emph{target} training samples \cite{Tax99}. The method tries to find the transformation (implicit in the kernel function) that maps the {\em target} data into a hypersphere. The proposed RBIG method and the SVDD method are conceptually similar due to their \emph{apparent} geometrical similarity. However, RBIG and SVDD represent two different approaches to the one-class classification problem: PDF estimation versus separation boundary estimation.
RBIG for one-class problems

may be naively seen as if test samples were transformed and classified as {\em target} if lying inside the sphere containing $1-\nu$ fraction of the learned Gaussian distribution. According to this interpretation, both methods reduce to the computation of spherical boundaries in different feature spaces. However, this is not true in the RBIG case: note that the value of the RBIG Jacobian is not the same at every location in the Gaussianized domain. Therefore, the optimal boundary to reject a $\nu$ fraction of the training data is not necessarily a sphere in the Gaussianized domain. In the case of the SVDD, though, by using an isotropic RBF kernel, all directions in the kernel feature spaces are treated in the same way.

\subsection{Relation to Deep Neural Networks}

RBIG is essentially an iterated sequence of
two operations: non-linear dimension-wise squashing functions and linear transforms.
Intuitively, these are the same processing blocks used in a feedforward neural network
(linear transform plus sigmoid-shaped function in each hidden layer).
Therefore, one could see each iteration as one
hidden layer processing of the data, and thus argue that
complex (highly non-Gaussian) tasks should require more hidden layers
(iterations). This view is in line with the field of {\em deep
learning} in neural networks \cite{Hinton06}, which consists of learning a
model with several layers of nonlinear mappings. The field is
very active nowadays because some tasks
are highly nonlinear and require accurate
design of processing steps of different complexity. Note, that
it may appear counterintuitive the fact that full Gaussianization
of a dataset is eventually achieved with a large enough number of iterations,
thus leading to overfitting in the case of a neural network
with such number of layers. Nevertheless, note that capacity control
also applies in RBIG: we have observed that early-stopping
criteria must be applied to allow good generalization properties.
In this setting, one can see early stopping in the Gaussianization method as a
form of model regularization. This is certainly an interesting
research line to be pursued in the future.

\vspace{0.5cm} 
{Finally, we would like to note that it does not escape our notice that the exploitation of the RBIG framework in the previous
contexts might eventually be helpful in designing new algorithms or helping understanding them
from different theoretical perspectives. This is of course out of the scope of this paper.}

\section{Experimental Results}
\label{applications}

This section shows the capabilities of the proposed RBIG methods in some illustrative examples. We start by experimentally analyzing the convergence of the method depending on the rotation matrix in a controlled toy dataset, and give useful criteria for early-stopping. Then, method's performance is illustrated for mutual information estimation, image synthesis, classification and denoising. In each application, results are compared to standard methods in the particular field.
A documented Matlab implementation is available at http://www.uv.es/vista/vistavalencia/RBIG.htm.

\subsection{Method convergence and early-stopping}
\label{Convergenceexample}

The RBIG method is analyzed here in terms of convergence rate and computational cost for
different rotations: orthonormal ICA, PCA and RND.
Synthetic data of varying dimensions ($d=2,\ldots,16$) was generated by first sampling from a uniform distribution hypercube and then applying a rotation
transform. This way we can compute the ground-truth negentropies of the initial distributions, and estimate the reduction in negentropies in every iteration by estimating the difference in marginal negentropies, cf. Eq. \eqref{prop1_explicit}. A total of $10000$ samples was used for the methods, and we show average and standard deviation results for $5$ independent random realizations.

\begin{figure}[t!]
\small
\begin{center}
\renewcommand{\tabcolsep}{0.05cm}
 \begin{tabular}{cccccc}
$k$ & 1 & 5 & 10 & 20 & 35 \\
RND &
 \includegraphics[scale=0.2]{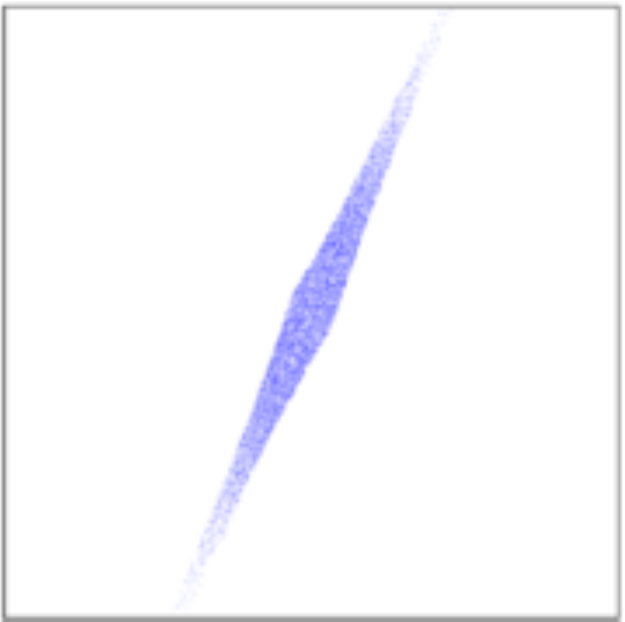}  &
 \includegraphics[scale=0.2]{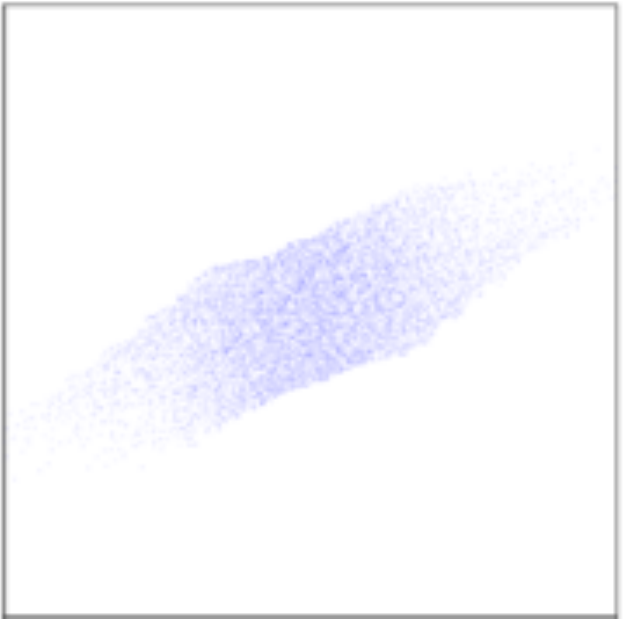}  &
 \includegraphics[scale=0.2]{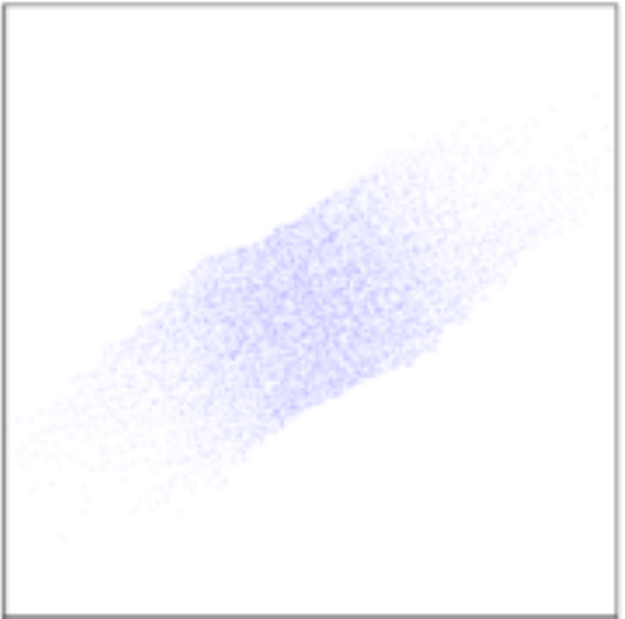}  &
 \includegraphics[scale=0.2]{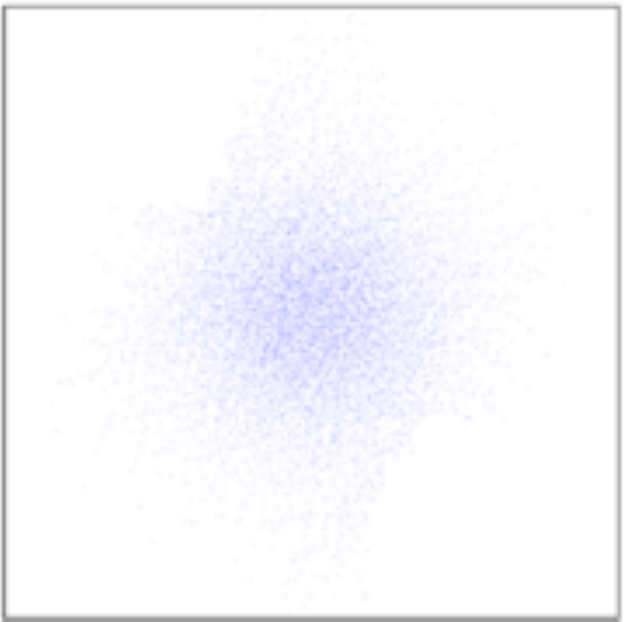}  &
 \includegraphics[scale=0.2]{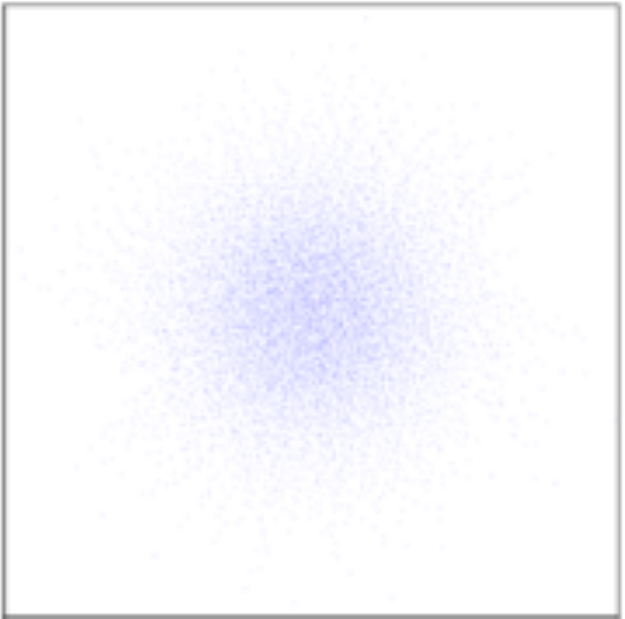}   \\
PCA &
 \includegraphics[scale=0.2]{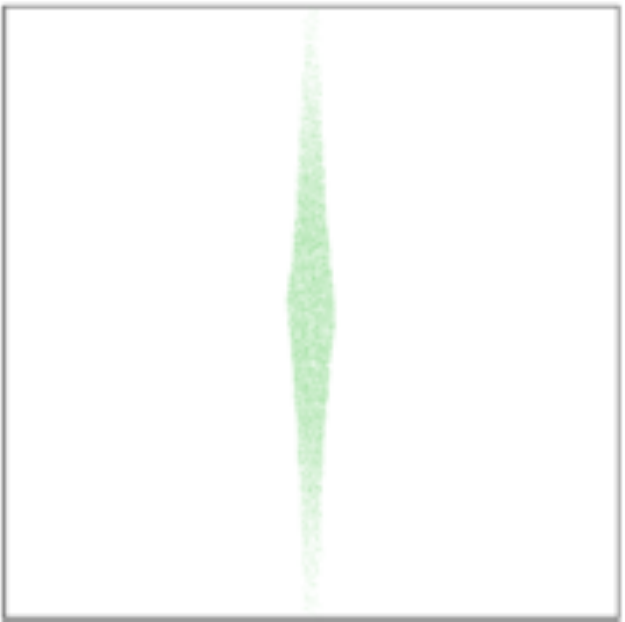}  &
 \includegraphics[scale=0.2]{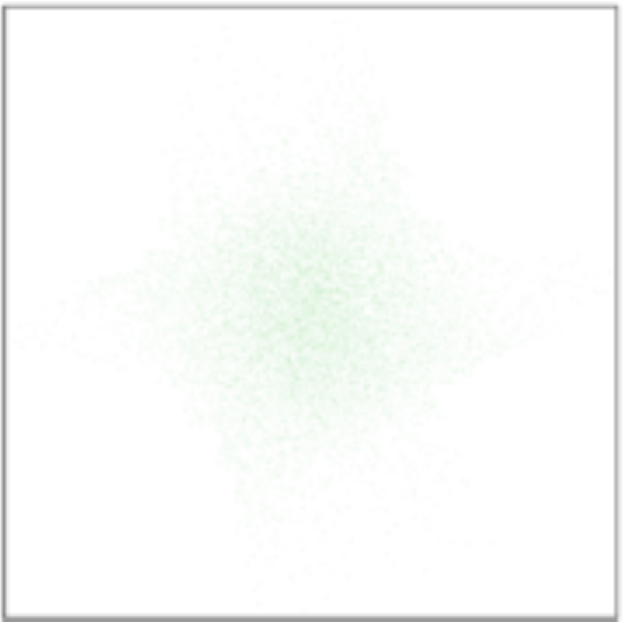}  &
 \includegraphics[scale=0.2]{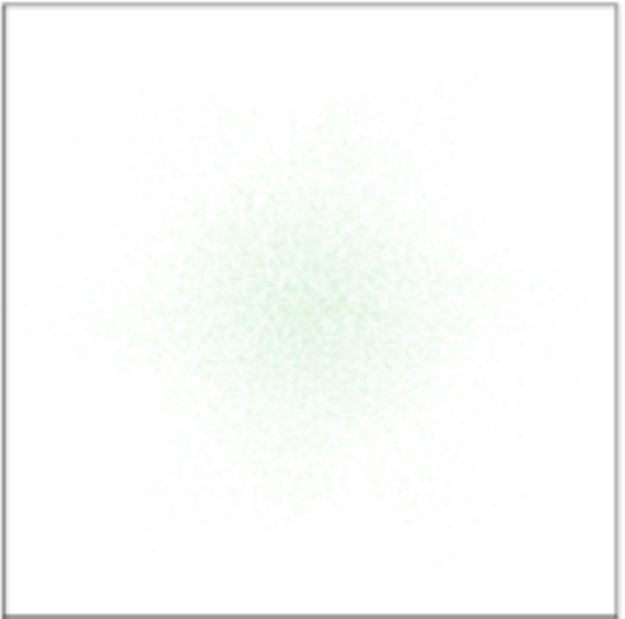}  &
 \includegraphics[scale=0.2]{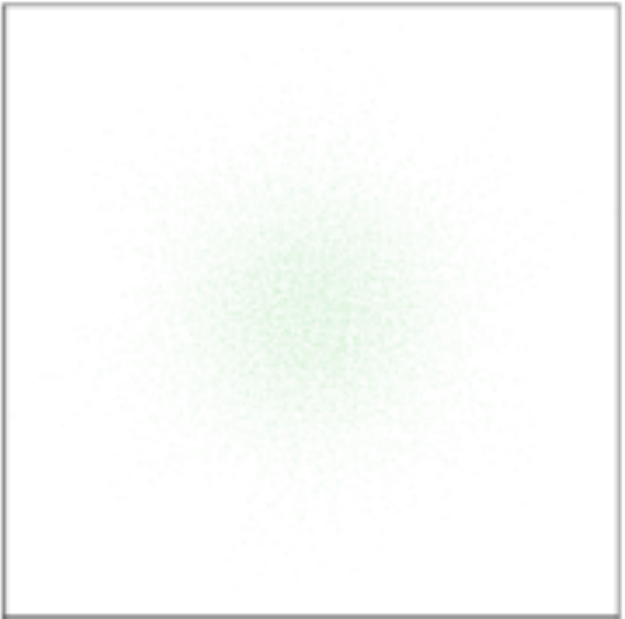}  &
 \includegraphics[scale=0.2]{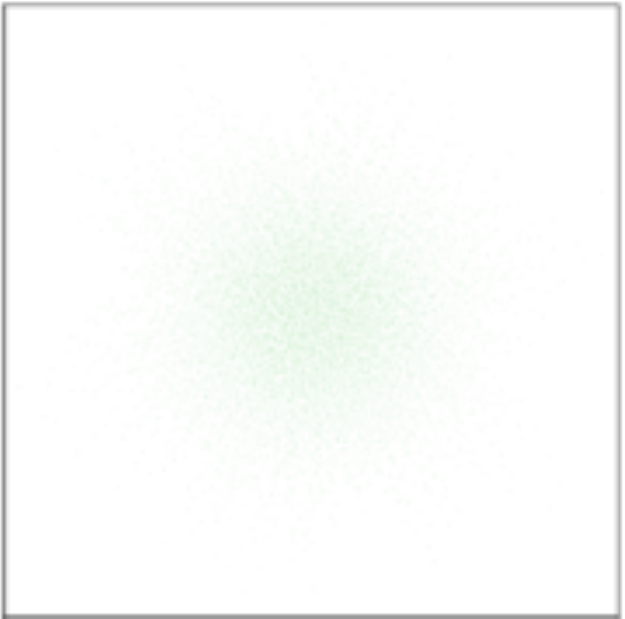}   \\
ICA &
 \includegraphics[scale=0.2]{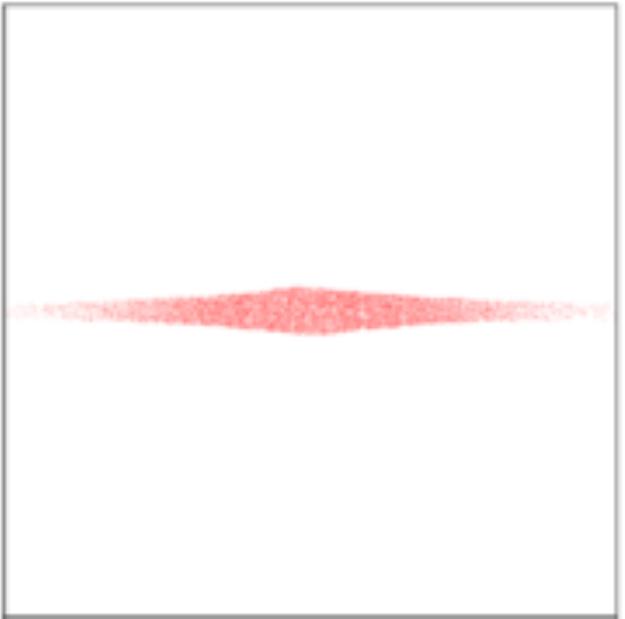}  &
 \includegraphics[scale=0.2]{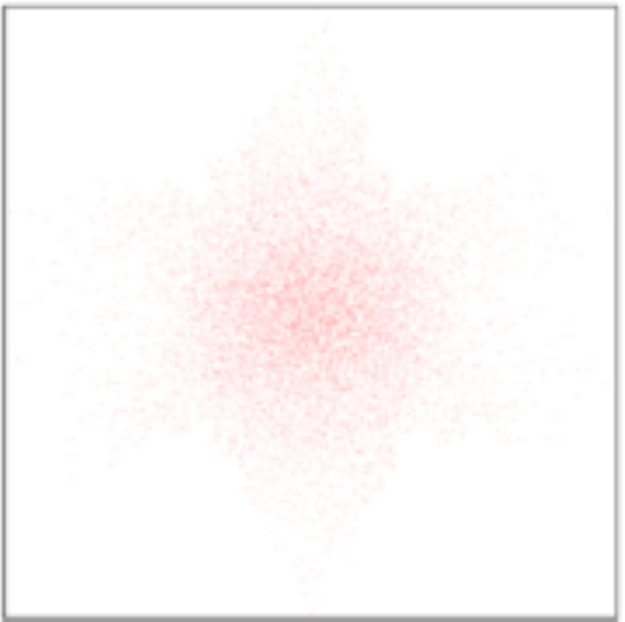}  &
 \includegraphics[scale=0.2]{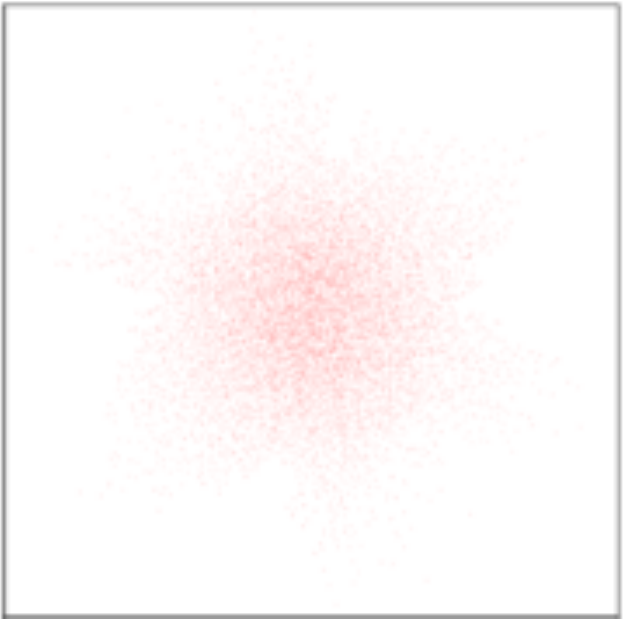}  &
 \includegraphics[scale=0.2]{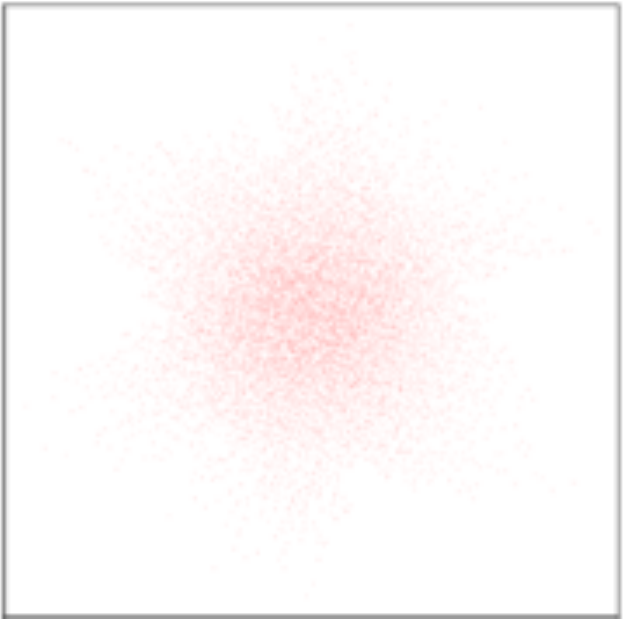}  &
 \includegraphics[scale=0.2]{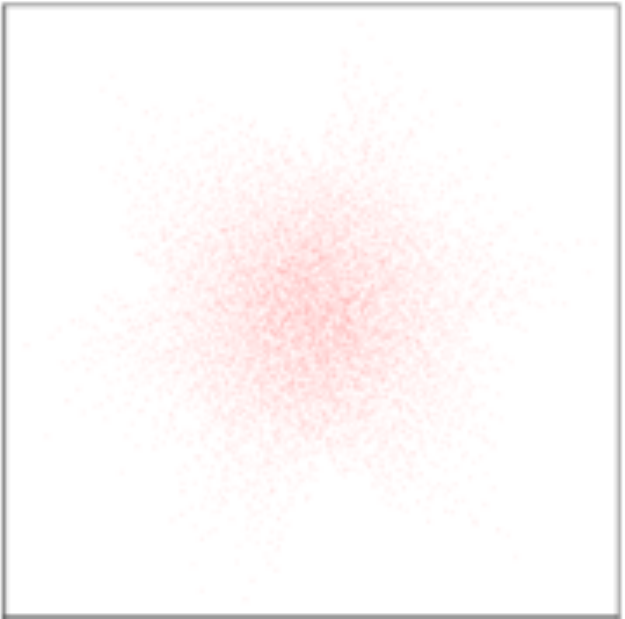}   \\
 \end{tabular}
\end{center}
 \caption{Scatter plots of a 2D data in different iterations for the considered rotation matrices: RND (top), PCA (middle) and ICA (bottom).}
\label{fig:convergence_scatters}
\end{figure}

Two-dimensional scatter plots in Figure \ref{fig:convergence_scatters} qualitatively show that different rotation matrices give rise to different solutions in each iteration but, after a sufficient number of iterations, all of them transform the data into a Gaussian independently of the rotation matrix.

\begin{figure}[t!]
\begin{center}
 \begin{tabular}{cc}
 \includegraphics[scale=0.45]{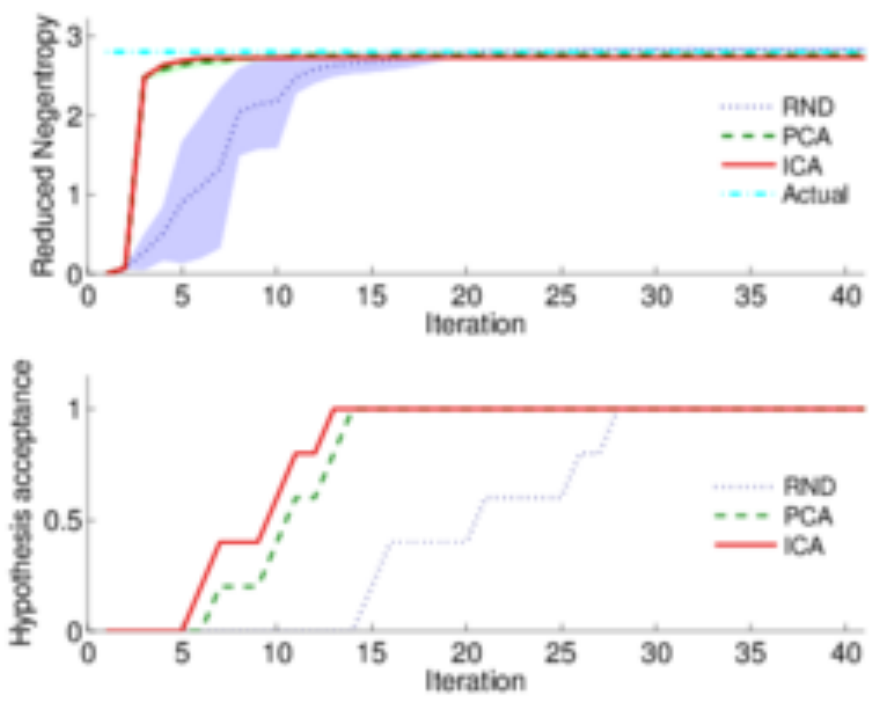} &
 \includegraphics[scale=0.45]{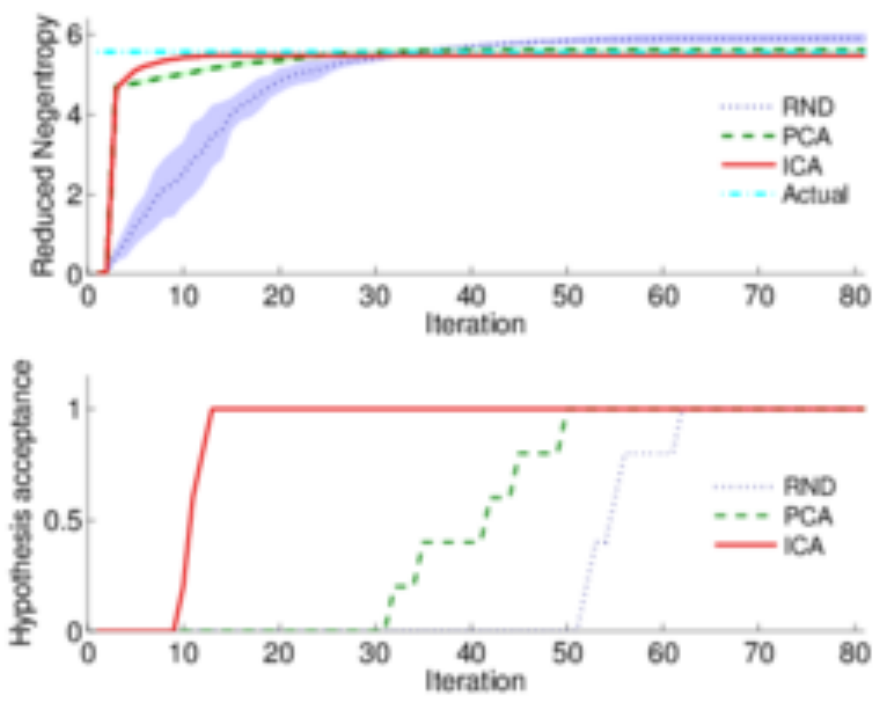}
 \end{tabular}
\end{center}
 \caption{Cumulative negentropy reduction (top) and multivariate Gaussian significance test (bottom) for each iteration in 2D (left) and 4D (right) dimensional synthetic problem. Average and standard deviation results from $5$ realizations is shown.}
\label{fig:convergence_MI}
\end{figure}

RBIG convergence rates are illustrated in Fig. \ref{fig:convergence_MI}.
Top plots show the negentropy reduction for the different rotations as a function of the number of iterations and data dimension. We also give the actual negentropy
estimated from the samples, is an univariate population estimate since Eq. \eqref{prop1} can be used. Successful convergence is obtained when the accumulated reduction in negentropy tends to the actual negentropy value (cyan line).
Discrepancies are due to the accumulation of computational errors in the negentropy reduction estimation in each iteration.

Bottom plots in Fig. \ref{fig:convergence_MI} give the result of the multivariate Gaussianity test in \cite{Szekely05}:
when the outcome of the test is $1$, it means accepting the hypothesis of multidimensional Gaussianity.
Several conclusions can be extracted: (1) the method converges to a multivariate Gaussian independently of the rotation matrix; (2) ICA requires a less number of iterations to converge, but it is closely followed by PCA; (3) random rotations take a higher number of iterations to converge and show high-variance in the earlier iterations; and
(4) convergence in cumulative negentropy is consistent with the parametric estimator in \cite{Szekely05} which, in turn, confirms the analysis in Table \ref{tablon}.

Despite the previous conclusions, and as pointed out before, in practical applications, it is not the length of the path to the Gaussian goal what matters, but the time required to complete this path. Table \ref{tab:conv_time} compares the number of iterations for appropriate convergence and the CPU time of $5$ realizations of RBIG with different matrix rotations (RND, PCA and ICA) in several dimensions. While, in general, CPU time results are obviously implementation dependent, note that results in Table \ref{tab:conv_time} are fairly consistent with the computational burden per iteration shown in Table \ref{tablon} since each ICA computation is an
iterative procedure itself which needs $m$ iterations.

\begin{table}[t!]
\scriptsize
\caption{Average ($\pm$ std. dev.) convergence results.}
\vspace{-0.5cm}
\begin{center}
\setlength{\tabcolsep}{2pt}
\begin{tabular}{c||c|c|c|c|c|c}
\hline
 & \multicolumn{2}{c|}{RND} & \multicolumn{2}{c|}{PCA} & \multicolumn{2}{c}{ICA} \\ \hline
Dim. & iterations & time [s] & iterations & time [s] & iterations & time [s]\\ \hline \hline
2 & 14 $\pm$ 3 & 0.01 $\pm$ 0.01 & 7 $\pm$ 3 & 0.005 $\pm$ 0.002 & 3 $\pm$ 1 & 6 $\pm$ 5 \\
4 & 44 $\pm$ 6 & 0.06 $\pm$ 0.01 & 33 $\pm$ 6 & 0.05 $\pm$ 0.01 & 11 $\pm$ 1 & 564 $\pm$ 223 \\
6 & 68 $\pm$ 7 & 0.17 $\pm$ 0.01 & 43 $\pm$ 12 & 0.1 $\pm$ 0 & 11 $\pm$ 2 & 966 $\pm$ 373 \\
8 & 92 $\pm$ 4 & 0.3 $\pm$ 0.1 & 54 $\pm$ 23 & 0.2 $\pm$ 0 & 16 $\pm$ 1 & 1905 $\pm$ 534 \\
10 & 106 $\pm$ 10 & 0.4 $\pm$ 0 & 58 $\pm$ 25 & 0.3 $\pm$ 0.1 & 19 $\pm$ 1 & 2774 $\pm$ 775 \\
12 & 118 $\pm$ 10 & 0.5 $\pm$ 0.2 & 44 $\pm$ 5 & 0.2 $\pm$ 0.1 & 21 $\pm$ 2 & 3619 $\pm$ 323 \\
14 & 130 $\pm$ 8 & 0.7 $\pm$ 0.1 & 52 $\pm$ 21 & 0.4 $\pm$ 0.1 & 19 $\pm$ 1 & 4296 $\pm$ 328 \\
16 & 139 $\pm$ 10 & 0.7 $\pm$ 0 & 73 $\pm$ 36 & 0.4 $\pm$ 0.2 & 22 $\pm$ 1 & 4603 $\pm$ 932 \\
\hline
\end{tabular}
\end{center}
\label{tab:conv_time}
\end{table}

The use of ICA rotations critically increases the convergence time. This effect is more noticeable as the dimension increases, thus making the use of ICA computationally unfeasible when the number of dimensions is moderate or high. The use of PCA in RBIG is consequently a good trade-off between Gaussianization error and computational cost if the number of iterations is properly chosen. An early-stopping criterion could be based on the evolution of the cumulative negentropy reduction, or of a multivariate test of Gaussianity such as the one used here \cite{Szekely05}. Both are sensible strategies for early-stopping. According to the observed performance, we restrict ourselves to the use of PCA as the rotation matrix in the experiments hereafter. Note that by using PCA, the algorithm might not converge in a singular situation (see Section \ref{On the rotation matrices}). However, we checked that such singular situation never happened by jointly using both criteria in each iteration.

\subsection{Multi-information Estimation}
\label{MI_measuring}

As previously shown, RBIG can be used to estimate the negentropy, and therefore could be used to compute multi-information ($I$) of high dimensional data (Eq. \eqref{negentropy-descomposition1}). Essentially, one learns the sequence of transforms to Gaussianize a given dataset, and the $I$ estimate reduces to compute the cumulative $\Delta I$ since, at convergence, full independence is supposedly achieved.
We illustrate the ability of RBIG in this context by estimating multi-information in three different synthetic distributions with known $I$: uniform distribution (UU), Gaussian distribution (GG), and a marginally composed exponential and Gaussian distribution (EG). An arbitrary rotation was applied in each case to obtain non-zero multi-information. In all cases, we used $10,000$ samples and repeated the experiments for $10$ realizations. Two kinds of experiments were performed:
\begin{itemize}
 \item A 2D experiment, where RBIG results can be compared to the results of naive (histogram-based) mutual information estimates (NE), and to previously reported 2D estimates such as the Rudy estimate (RE) \cite{Moddemeijer89} (see Table \ref{tab:2d_MI_results}).
 \item A set of $d$-dimensional experiments, where RBIG results are compared to actual values (see Table IV).
\end{itemize}
Table \ref{tab:2d_MI_results} shows the results (in bits) for the mutual information estimation
in the 2D experiment to standard approaches. The ground-truth result is also given for comparison purposes.

\begin{table}[t!]
\small
\caption{Average ($\pm$ std. dev.) multi-information (in bits) for the different estimators in 2D problems.}
\vspace{-0.25cm}
\begin{center}
\setlength{\tabcolsep}{4pt}
\begin{tabular}{c||c|c|c}
\hline
DIST & EG & GG & UU \\ \hline
RBIG & 0.49 $\pm$ 0.01 & 1.38 $\pm$ 0.004 & 0.36 $\pm$ 0.03 \\
NE & 0.35 $\pm$ 0.02 & 1.35 $\pm$ 0.006 & 0.39 $\pm$ 0.002 \\
RE & 0.32 $\pm$ 0.01 & 1.29 $\pm$ 0.004 & 0.30 $\pm$ 0.002 \\
\hline
Actual & 0.51 & 1.38 & 0.45  \\
\hline
\end{tabular}
\end{center}
\label{tab:2d_MI_results}
\end{table}
For Gaussian and exponential-Gaussian data distributions, RBIG outperforms the rest of methods, but when data are marginally uniform, NE yields better estimates.
Table IV extends the previous results to multidimensional cases, and compares RBIG to the actual $I$. Good results are obtained in all cases. Absolute errors slightly increase with data dimensionality.

\begin{table}[t!]
\scriptsize
\begin{center}
\caption{Multi-information (in bits) with RBIG in different $d$-dimensional problems.}
\setlength{\tabcolsep}{4pt}
\begin{tabular}{c|c|c|c|c|c|c}
\hline
Dim.               & \multicolumn{2}{c|}{EG} & \multicolumn{2}{c|}{GG} & \multicolumn{2}{c}{UU} \\
\cline{2-7}
$d$ & RBIG & Actual & RBIG & Actual & RBIG & Actual\\
\hline
3 & 1.12 $\pm$ 0.03 & 1.07  & 1.91 $\pm$ 0.01 & 1.9    & 1.6 $\pm$ 0.1 & 1.6   \\
4 & 5 $\pm$ 0.1 & 5.04  & 1.88 $\pm$ 0.02 & 1.86   & 2.2 $\pm$ 0.1 & 2.2   \\
5 & 4.7 $\pm$ 0.1 & 4.82  & 1.77 $\pm$ 0.02 & 1.75   & 2.7 $\pm$ 0.1 & 2.73   \\
6 & 7.8 $\pm$ 0.1 & 7.9  & 2.11 $\pm$ 0.01 & 2.08   & 3.5 $\pm$ 0.1 & 3.72   \\
7 & 6.2 $\pm$ 0.1 & 6.33  & 2.68 $\pm$ 0.03 & 2.65   & 3.6 $\pm$ 0.1 & 3.92   \\
8 & 8.1 $\pm$ 0.1 & 8.19  & 2.72 $\pm$ 0.02 & 2.68   & 4.1 $\pm$ 0.1 & 4.29   \\
9 & 9.5 $\pm$ 0.1 & 9.6  & 3.22 $\pm$ 0.02 & 3.18   & 5.3 $\pm$ 0.1 & 5.69   \\
10 & 12.7 $\pm$ 0.1 & 13.3  & 3.45 $\pm$ 0.03 & 3.4   & 5.8 $\pm$ 0.2 & 6.24   \\
\hline
\end{tabular}
\end{center}
\label{cojones}
\end{table}

\subsection{Data Synthesis}

RBIG obtains an invertible Gaussianization transform that can be used to generate (or synthesize) samples. The approach is simple: the transform ${\mathcal G}$ is {\em learned} from the available training data, and then synthesized samples are obtained from random Gaussian samples in the transformed domain
inverted back to the original domain using ${\mathcal G}^{-1}$.
Two examples are given here to illustrate the capabilities of the method.

\subsubsection{Toy data}

\begin{figure}[t!]
\small
\begin{center}
\begin{tabular}{ccc}
{\bf Original data} & {\bf Gaussianized data} & {\bf Synthesized data} \\ 
\includegraphics[height=2.5cm,width=2.5cm]{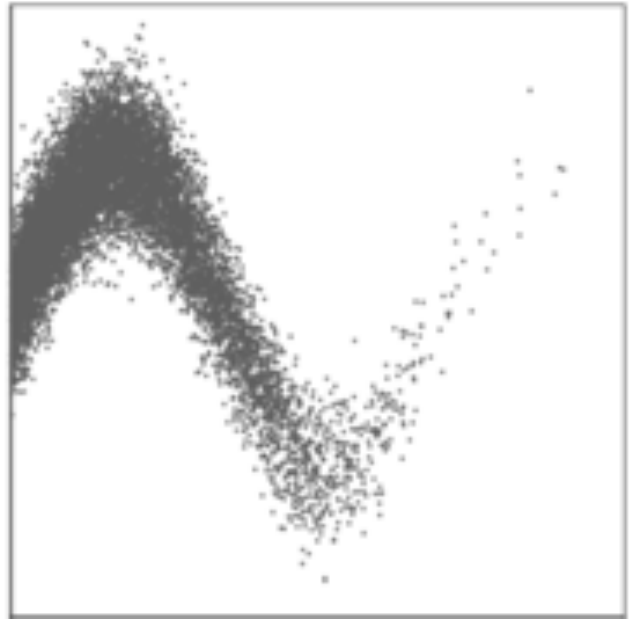}   &
\includegraphics[height=2.5cm,width=2.5cm]{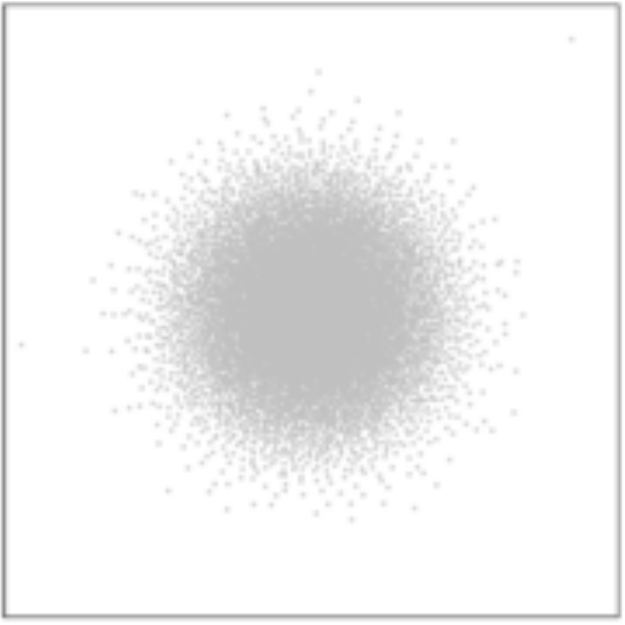}   &
\includegraphics[height=2.5cm,width=2.5cm]{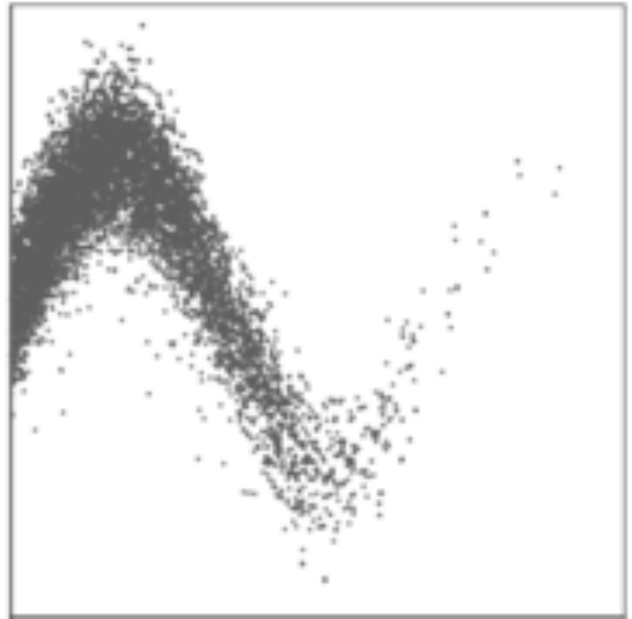}   \\[2mm]
\includegraphics[height=2.5cm,width=2.5cm]{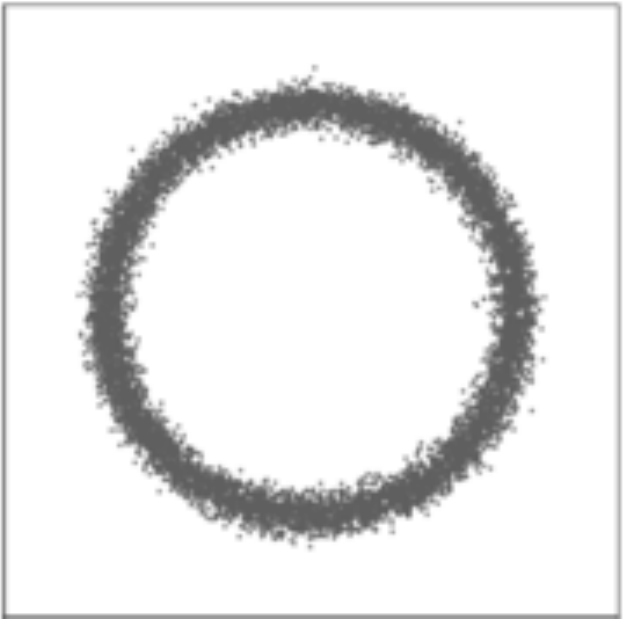}   &
\includegraphics[height=2.5cm,width=2.5cm]{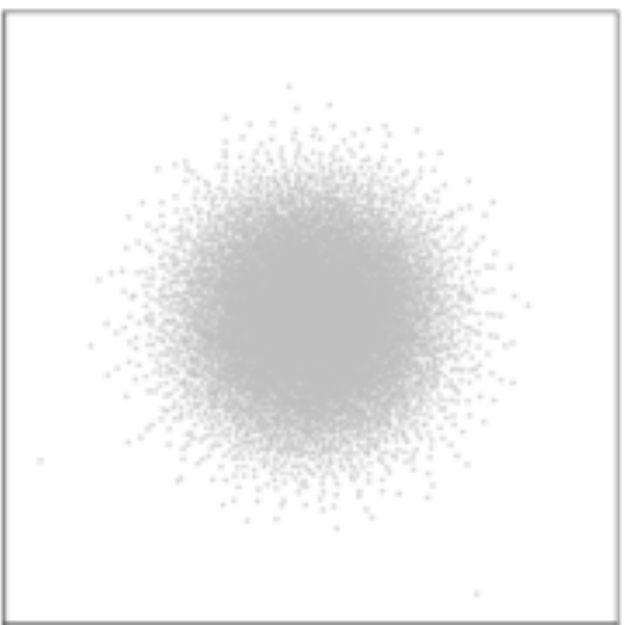}   &
\includegraphics[height=2.5cm,width=2.5cm]{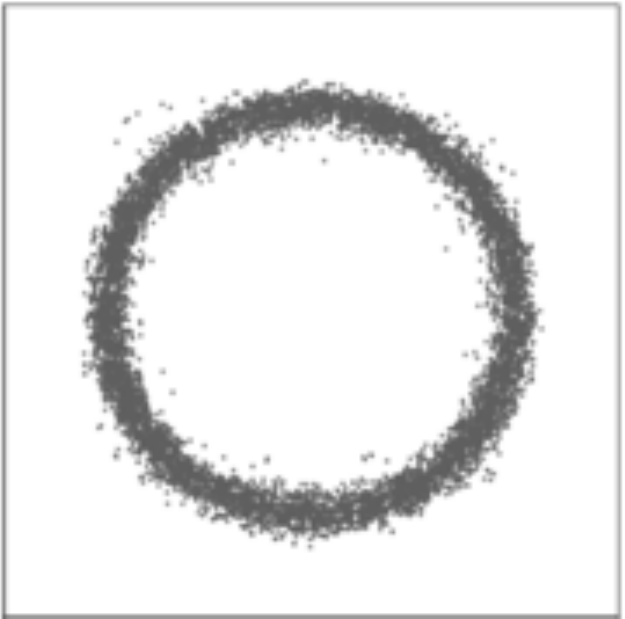}   \\[2mm]
\includegraphics[height=2.5cm,width=2.5cm]{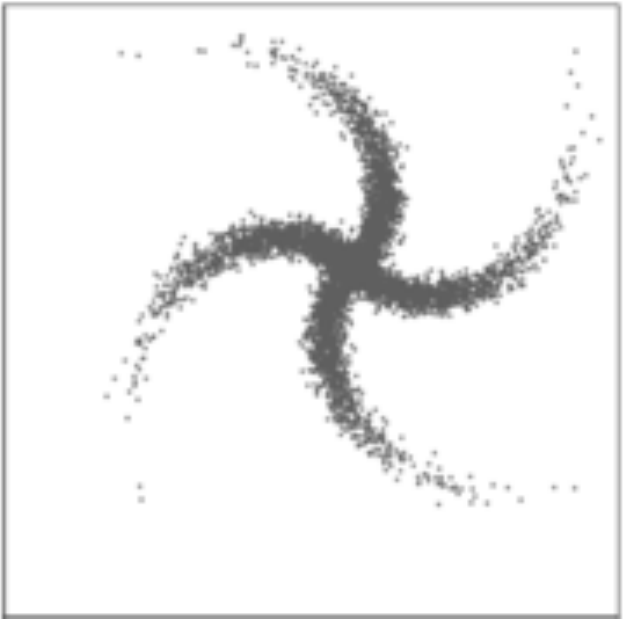}   &
\includegraphics[height=2.5cm,width=2.5cm]{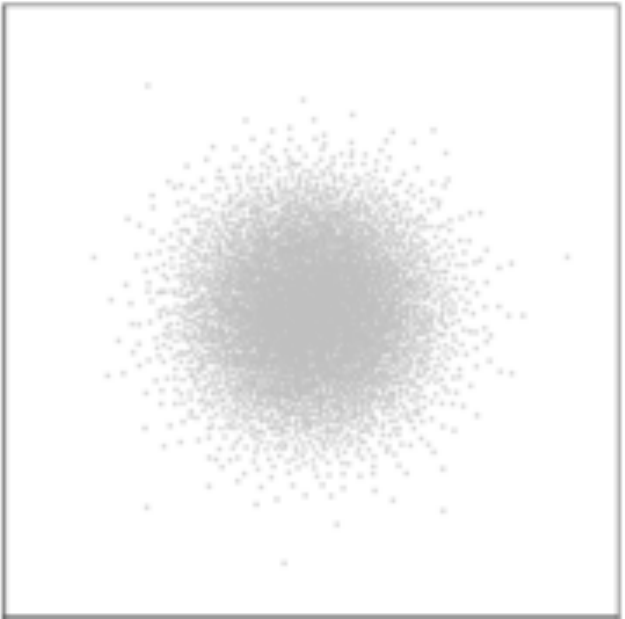}   &
\includegraphics[height=2.5cm,width=2.5cm]{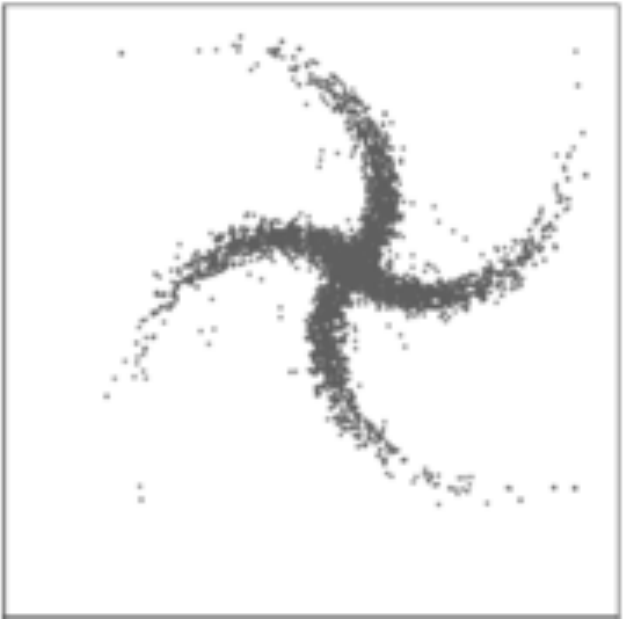}   \\[2mm]
 \end{tabular}
\end{center}
\vspace{-0.25cm}
 \caption{Toy data examples synthesized using RBIG.}
\label{fig:synthesis_2d}
\end{figure}

Figure \ref{fig:synthesis_2d} shows examples of 2D non-Gaussian distributions (left column) transformed into a Gaussian (center column). The right column was obtained sampling data from a zero mean unit covariance Gaussian and inverting back the transform.
This example visually illustrates that synthesized data approximately follow the original PDF.

\subsubsection{Face synthesis}

\begin{figure*}[t!]
\begin{center}
 \begin{tabular}{lcccccccccc}
{\bf Original} &
 \includegraphics[scale=0.205]{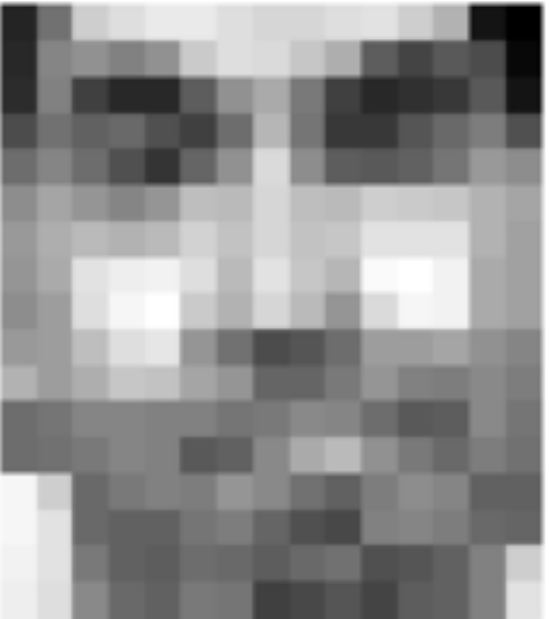} & 
 \includegraphics[scale=0.205]{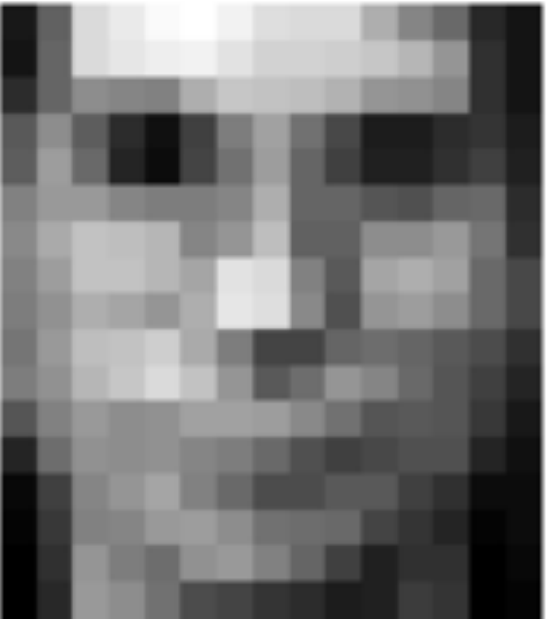} & 
 \includegraphics[scale=0.205]{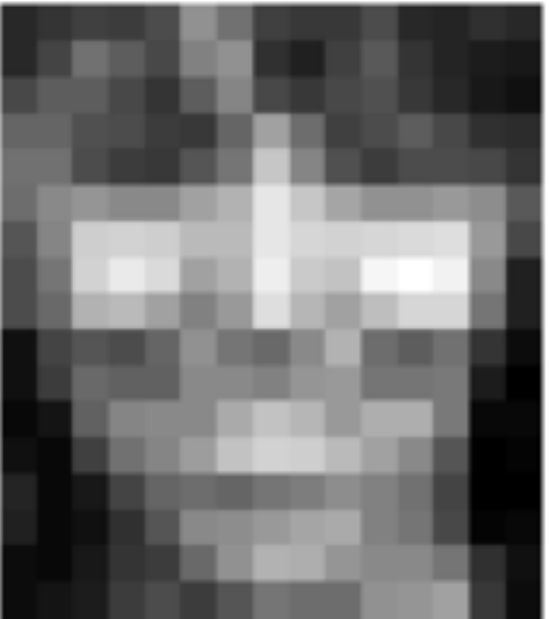} & 
 \includegraphics[scale=0.205]{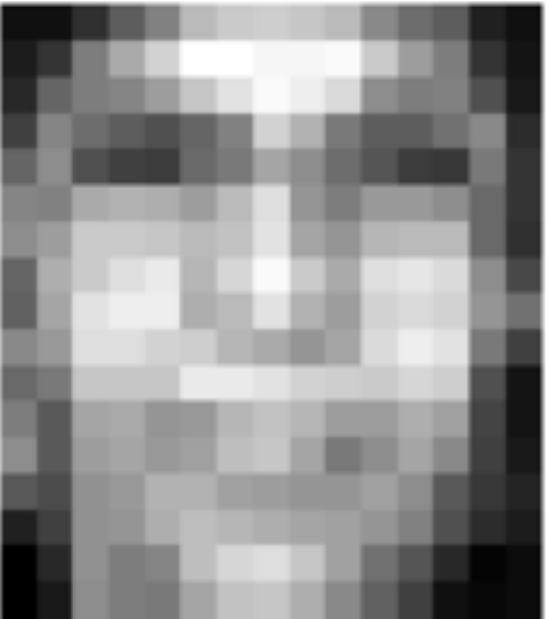} & 
 \includegraphics[scale=0.205]{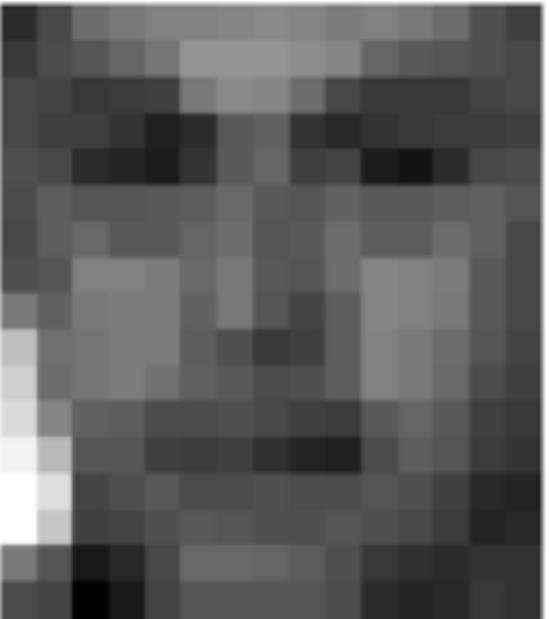} & 
 \includegraphics[scale=0.205]{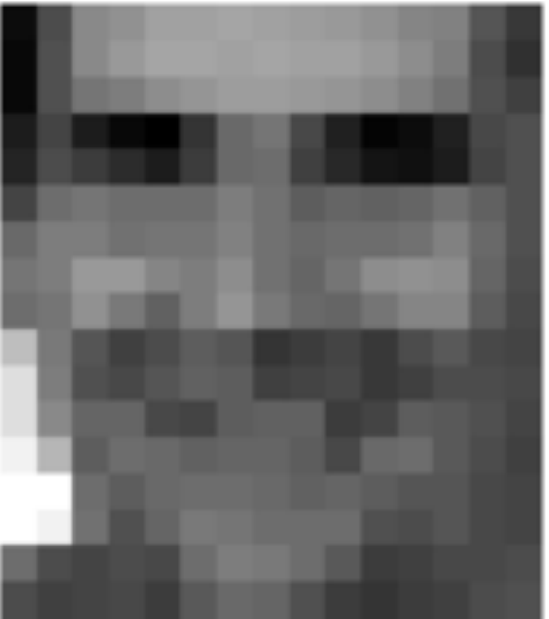} & 
 \includegraphics[scale=0.205]{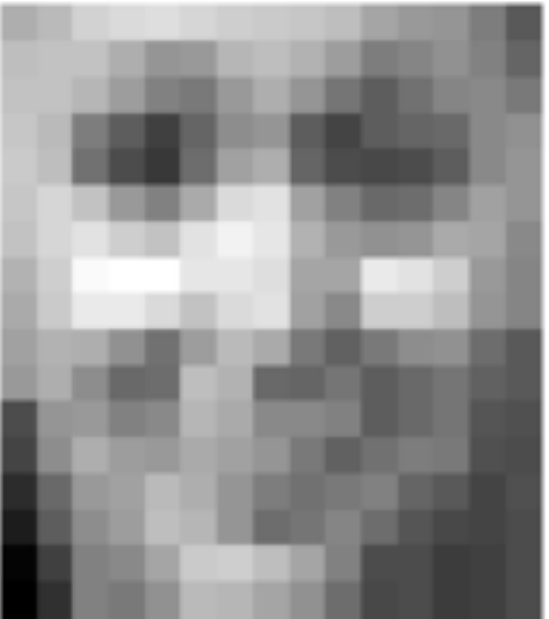} & 
 \includegraphics[scale=0.205]{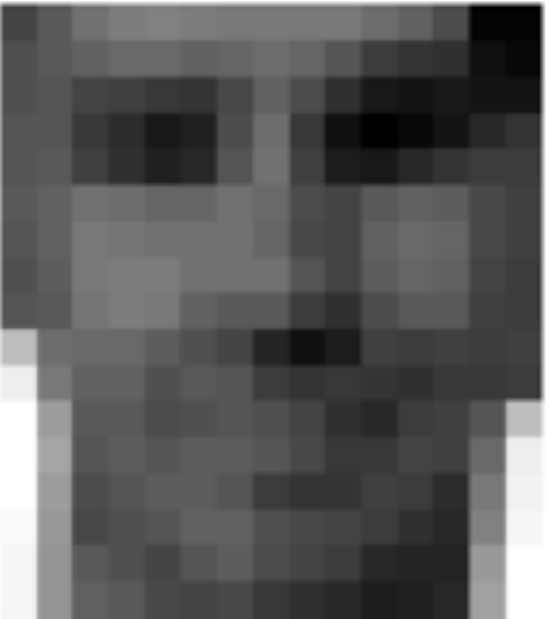} & 
 \includegraphics[scale=0.205]{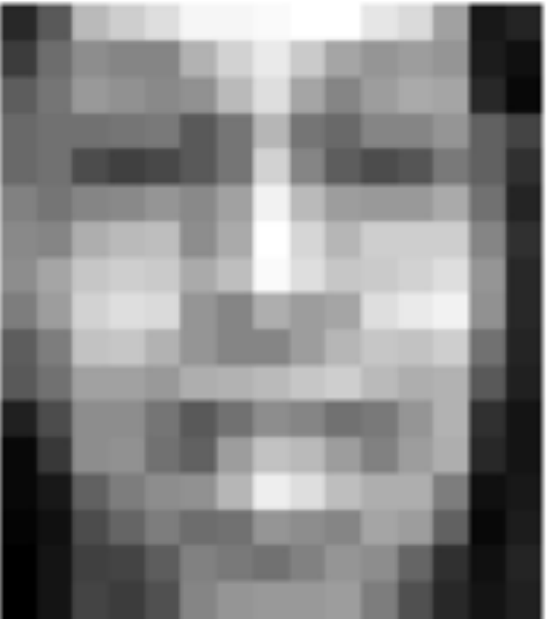} & 
 \includegraphics[scale=0.205]{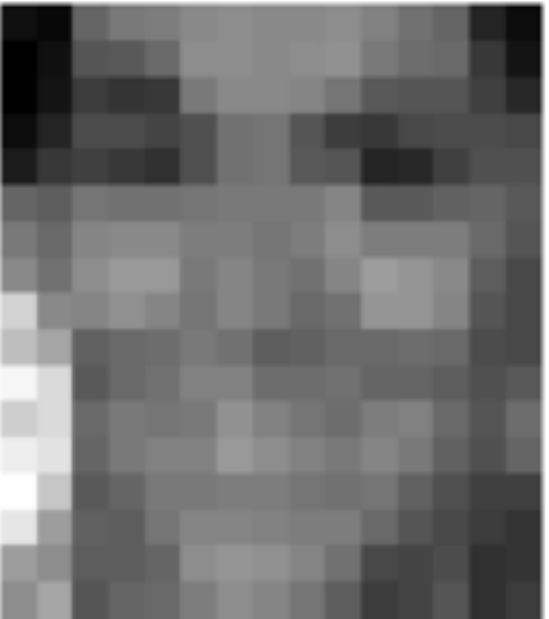} 
 \\
{\bf RG} &
 \includegraphics[scale=0.205]{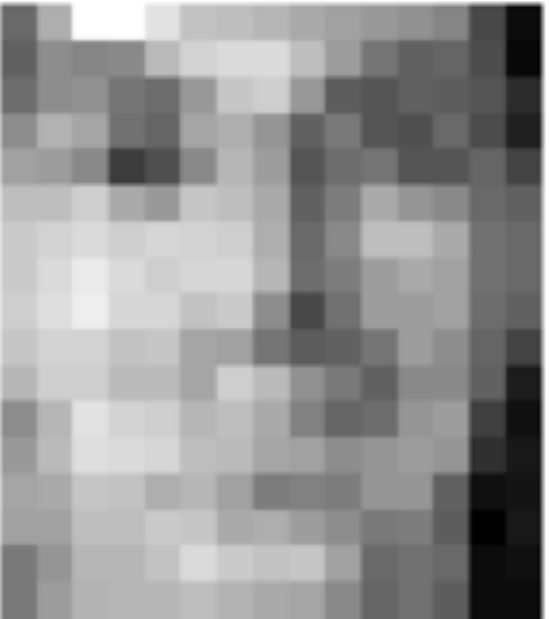} & 
 \includegraphics[scale=0.205]{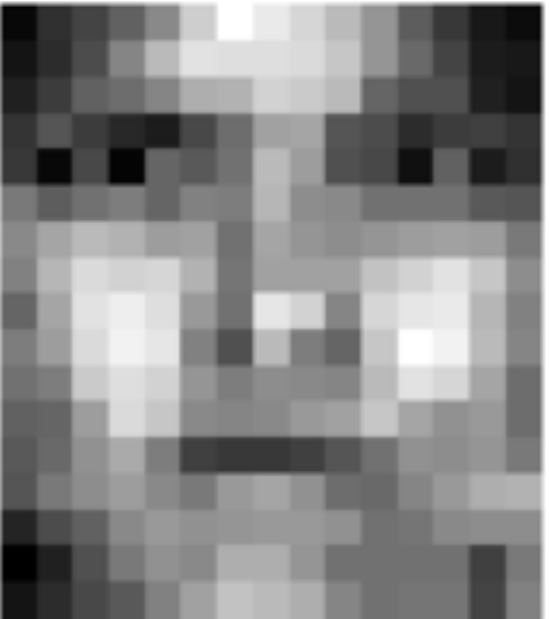} & 
 \includegraphics[scale=0.205]{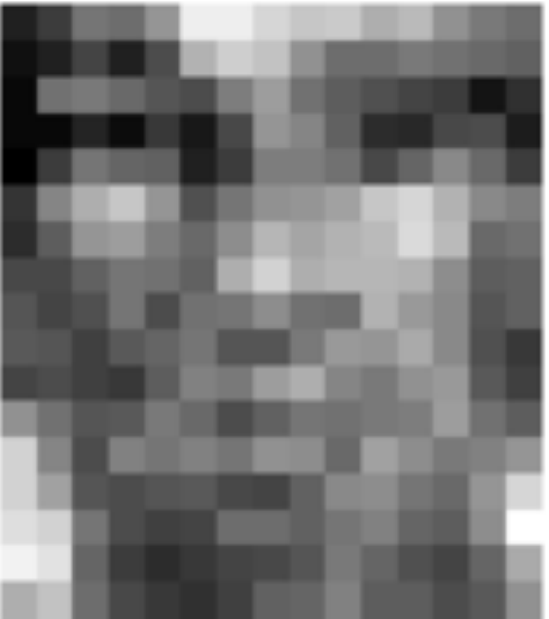} & 
 \includegraphics[scale=0.205]{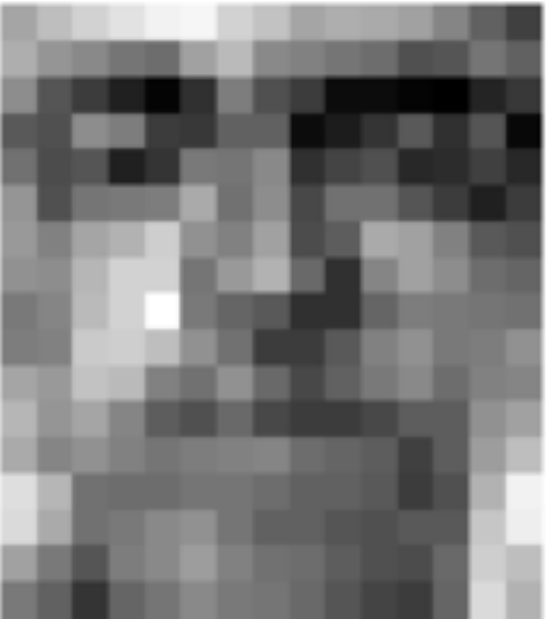} & 
 \includegraphics[scale=0.205]{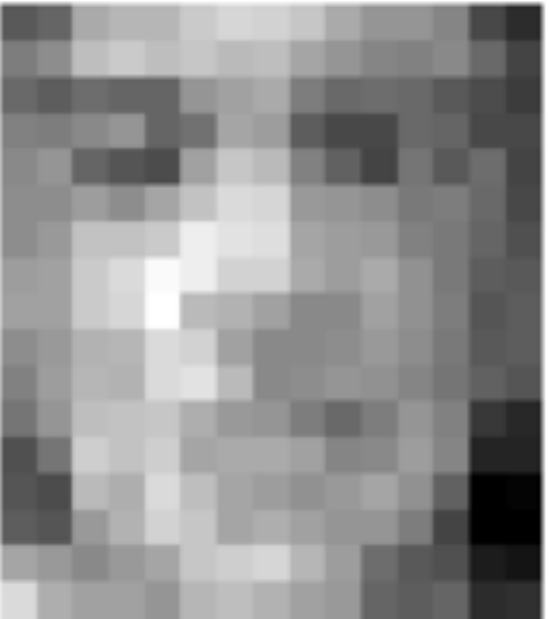} & 
 \includegraphics[scale=0.205]{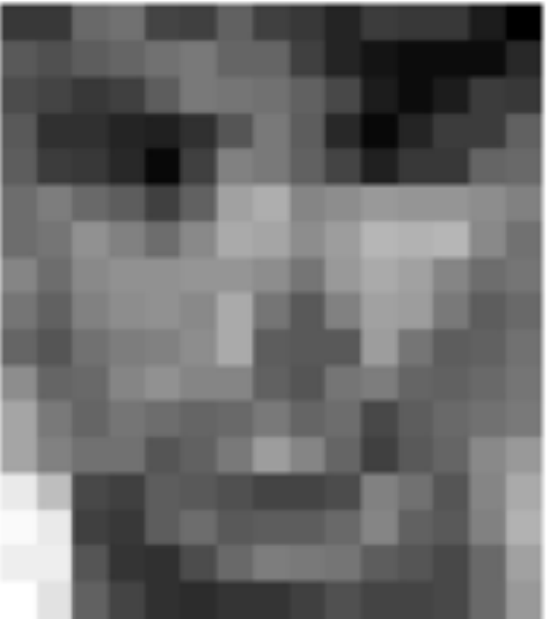} & 
 \includegraphics[scale=0.205]{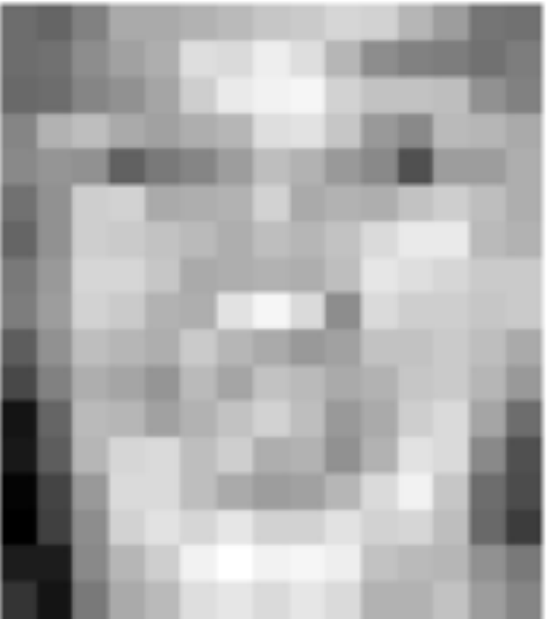} & 
 \includegraphics[scale=0.205]{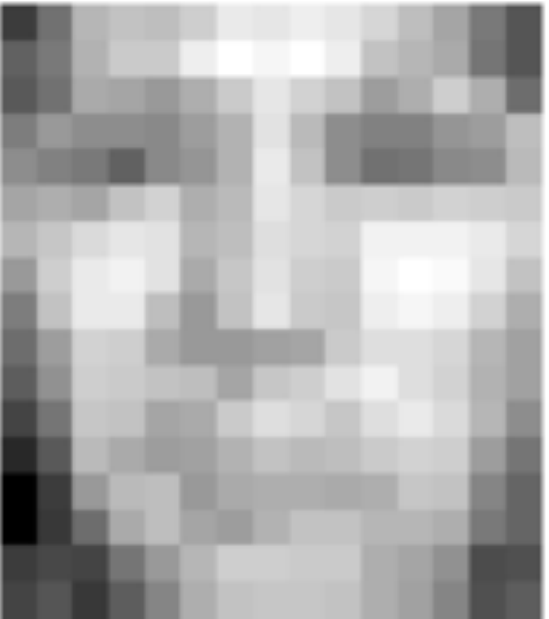} & 
 \includegraphics[scale=0.205]{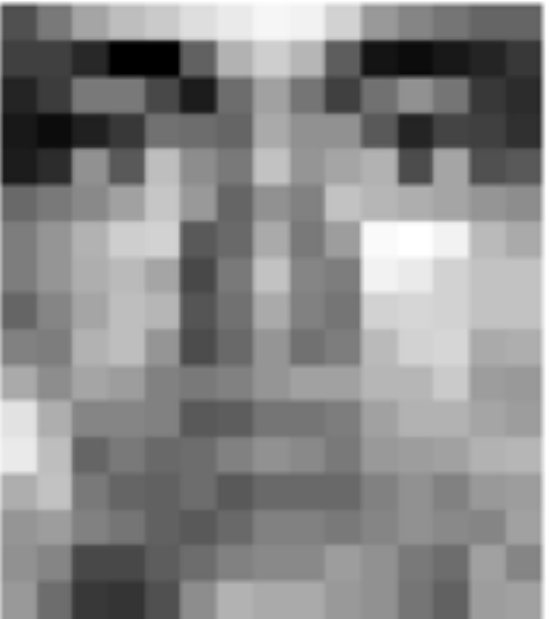} & 
 \includegraphics[scale=0.205]{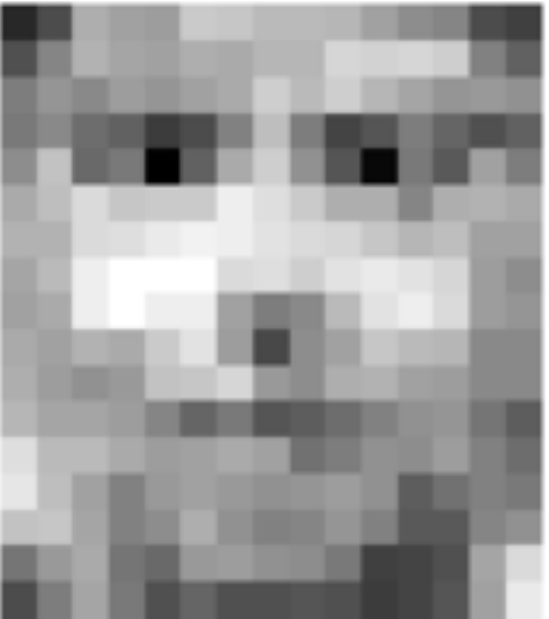} 
\\
{\bf RBIG} &
 \includegraphics[scale=0.205]{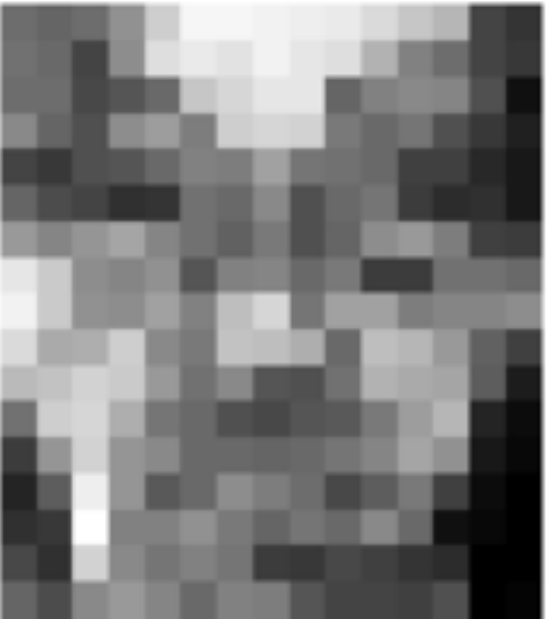} & 
 \includegraphics[scale=0.205]{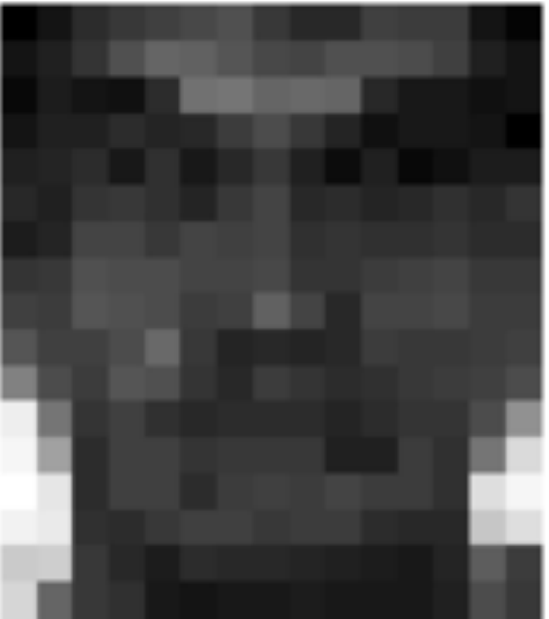} & 
 \includegraphics[scale=0.205]{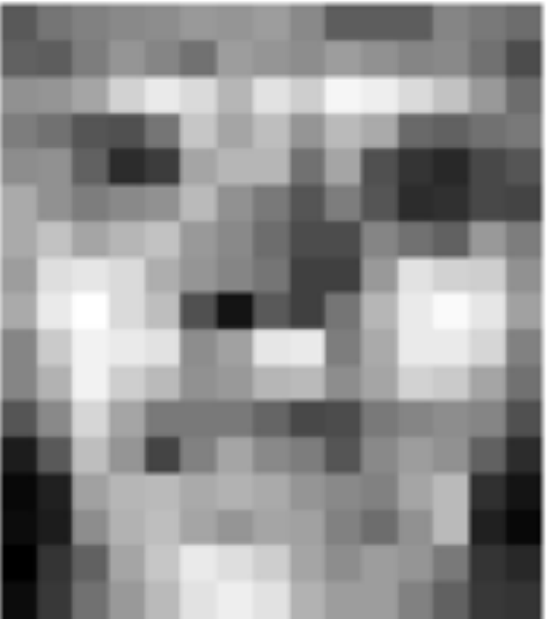} & 
 \includegraphics[scale=0.205]{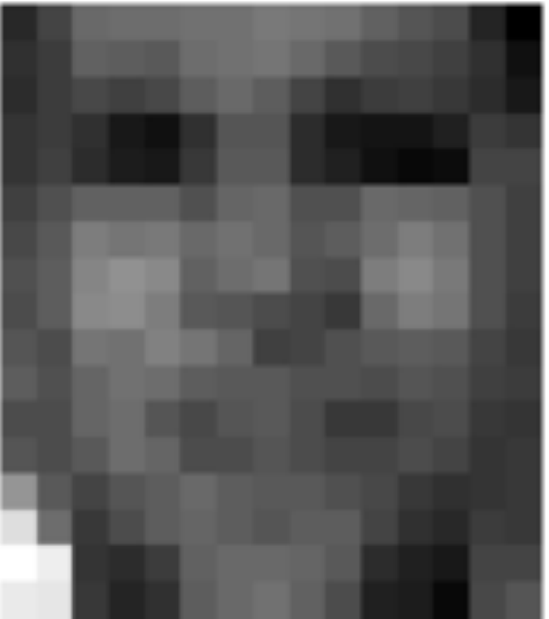} & 
 \includegraphics[scale=0.205]{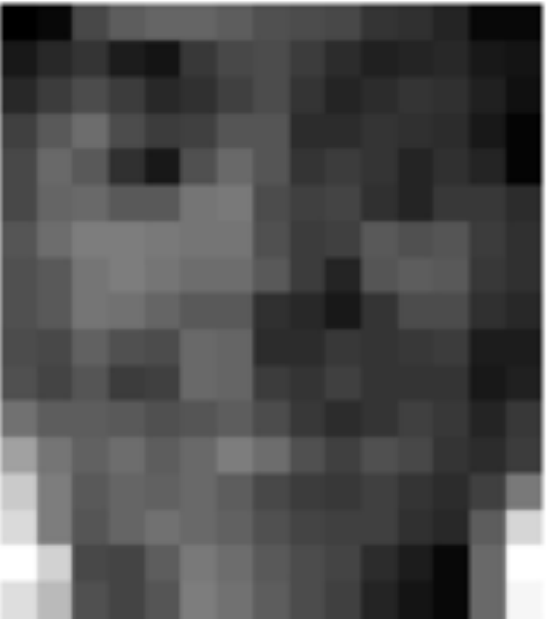} & 
 \includegraphics[scale=0.205]{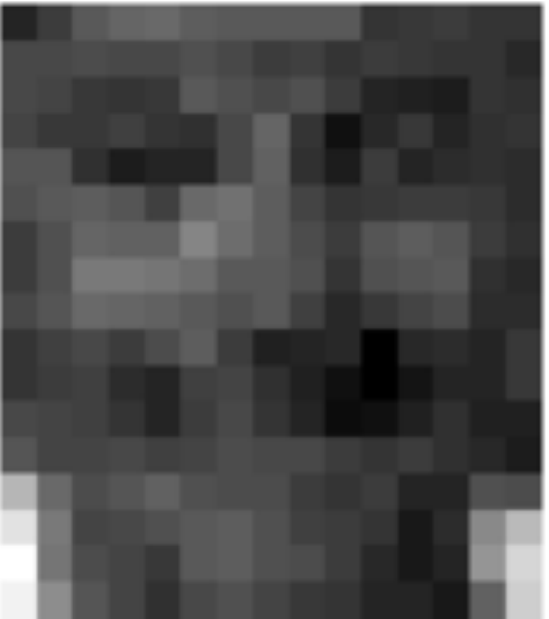} & 
 \includegraphics[scale=0.205]{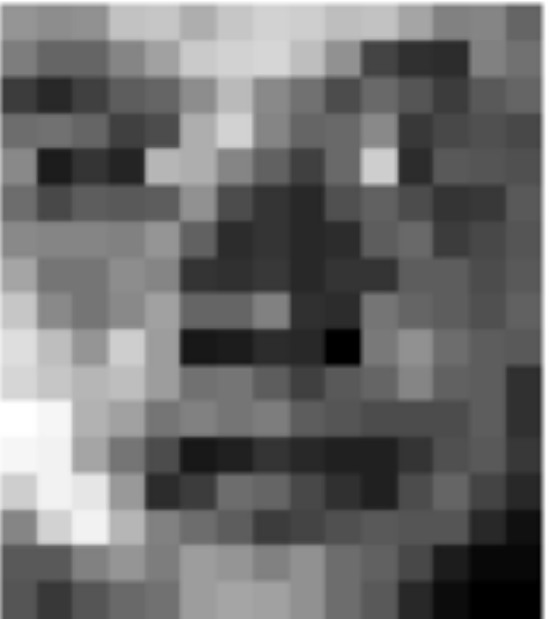} & 
 \includegraphics[scale=0.205]{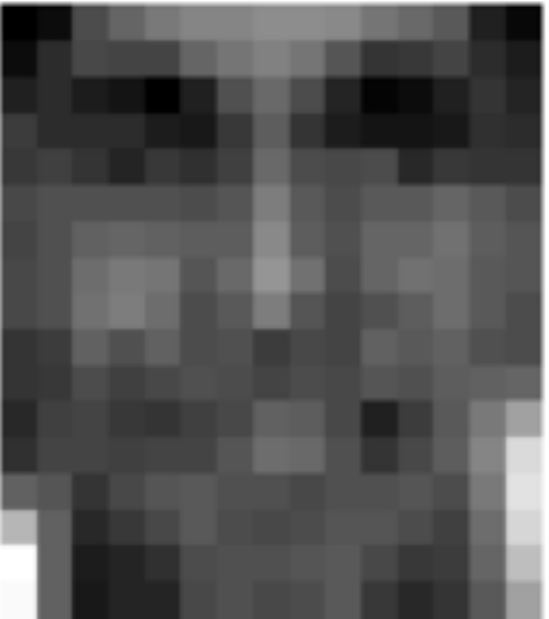} & 
 \includegraphics[scale=0.205]{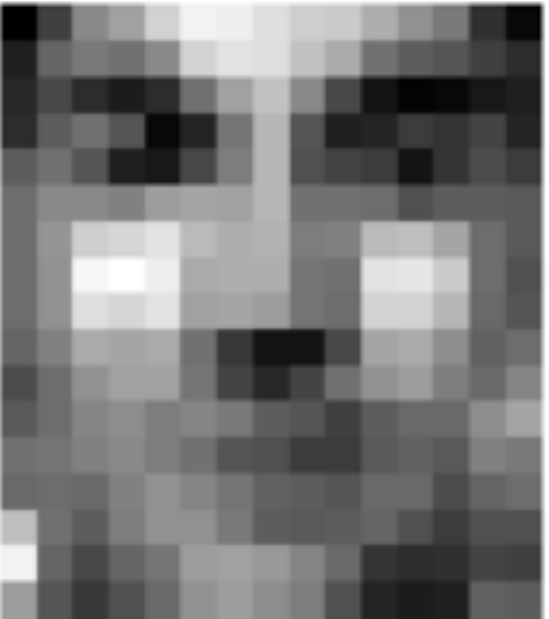} & 
 \includegraphics[scale=0.205]{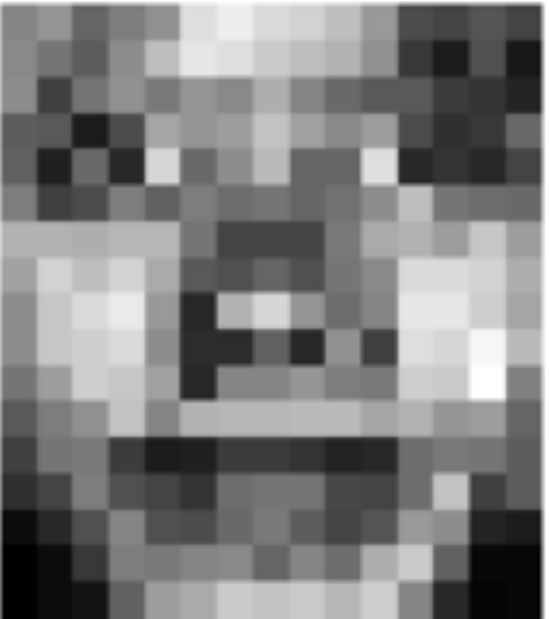}
 \end{tabular}
\end{center}
 \caption{Example of real (top) and synthesized faces with RG (middle) and RBIG (bottom).}
\label{fig:synth_faces}
\end{figure*}

In this experiment, $2,500$ face images were extracted from \cite{Georghiades01}, eye-centered, cropped to have the same dimensions, mean and variance adjusted, and resized to $17 \times 15$ pixels. Images were then reshaped to $255$-dimensional vectors, and Gaussianized with RG and RBIG.
Figure \ref{fig:synth_faces} shows illustrative examples of original and synthesized faces with RG and RBIG.

Note that both methods achieve good visual qualitative performance. In order to assess performance quantitatively, we compared $200$ actual and synthesized images using the inner product as a measure of local similarity. We averaged this similarity measure over $300$ realizations and show the histograms for RG and RBIG.
Results suggest that the distribution of the samples generated with RBIG is more realistic (similar to the original dataset) than the obtained with RG.

\subsection{One-class Classification}

In this experiment, we assess the performance of the RBIG method as one-class classifier. Performace is illustrated in the challenging problem of detecting urban areas from multispectral and SAR images. The ground-truth data for the images used in this section were collected in the Urban Expansion Monitoring (UrbEx) ESA-ESRIN DUP project\footnote{{http://dup.esrin.esa.int/ionia/projects/summaryp30.asp}} \cite{GomezChovaPRL06}.
The considered test sites were the cities of Rome and Naples, Italy, for two acquisitions dates (1995 and 1999). The available features were the seven Landsat bands, two SAR backscattering intensities (0--35 days), and the SAR interferometric coherence. We also used a spatial version of the coherence specially designed to increase the urban areas discrimination \cite{GomezChovaPRL06}. After this preprocessing, all features were stacked at a pixel level, and each feature was standardized.

\begin{figure}[t!]
\centerline{\includegraphics[width=7cm]{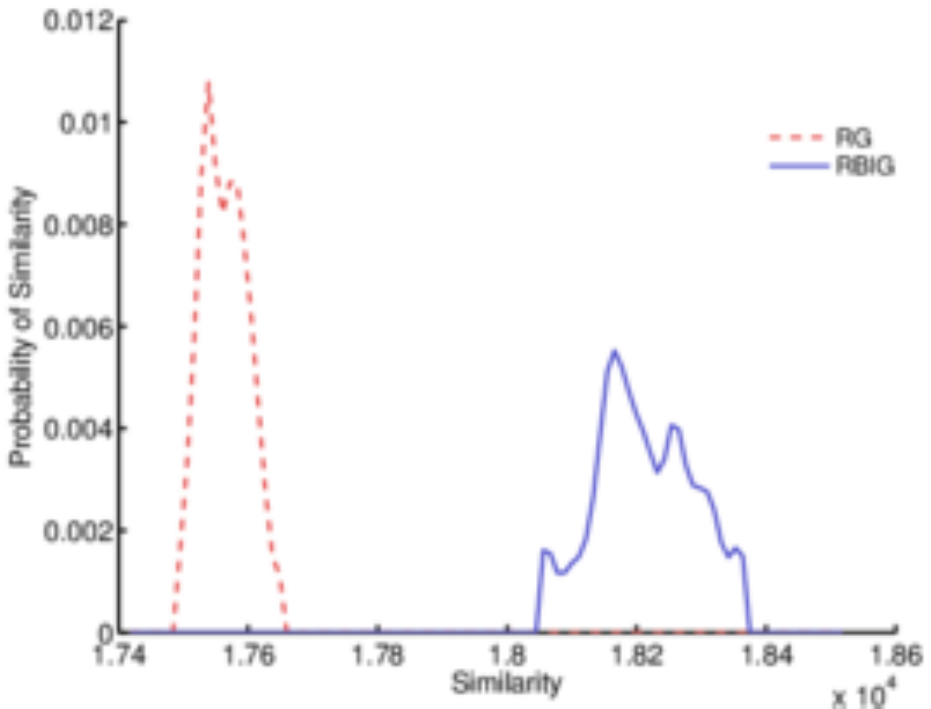}}
 \caption{Histogram of the similarity (inner product) between the distribution of original and synthesized face images for $300$ realizations. For reference, the average image energy (average inner product) in the original set is $1.81\cdot10^4$.}
\label{fig:synt_faces_measures}
\end{figure}

We compared the RBIG classifier based on the estimated PDF for urban areas with the SVDD classifier \cite{Tax99}. We used the RBF kernel for the SVDD whose width was varied in the range $\sigma\in[10^{-2},\ldots, 10^2]$. The fraction rejection parameter was varied in $\nu\in[10^{-2}, 0.5]$ for both methods. The optimal parameters were selected through $3$-fold cross-validation in the training set optimizing the $\kappa$ statistic \cite{Cohen60}. Training sets of different size for the target class were used in the range $[500, 2500]$. We assumed a scarce knowledge of the non-target class: $10$ outlier examples were used in all cases. The test set was constituted by $10,000$ pixels of each considered image. Training and test samples were randomly taken from the whole spatial extent of each image. The experiment was repeated for $10$ different random realizations in the three considered test sites.

\begin{figure}[t!]
\begin{center}
\begin{tabular}{cc}
\includegraphics[width=0.45\columnwidth]{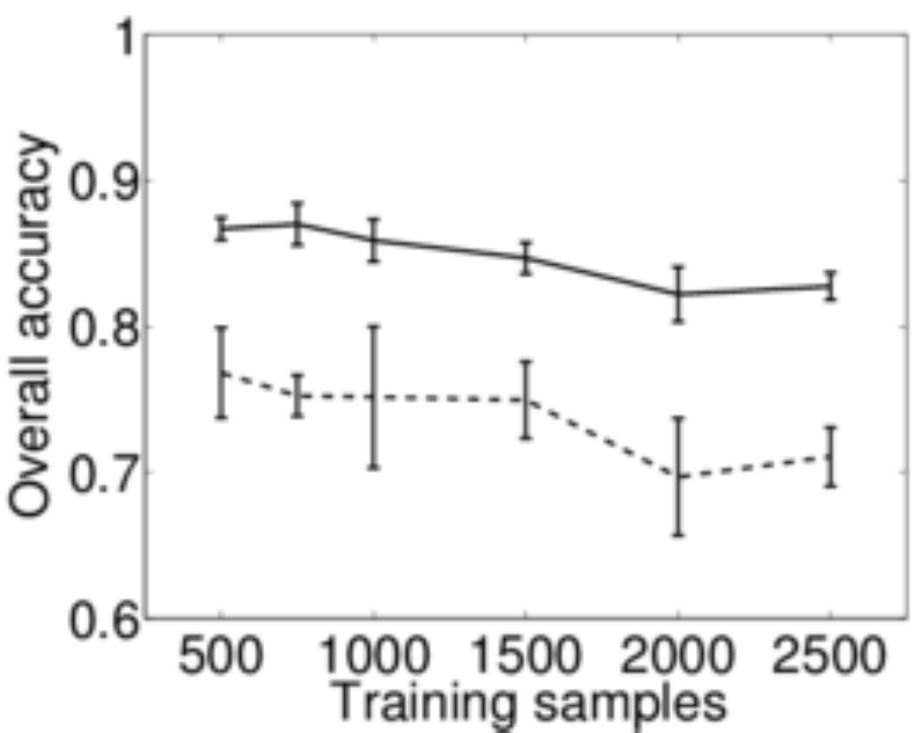} & \includegraphics[width=0.45\columnwidth]{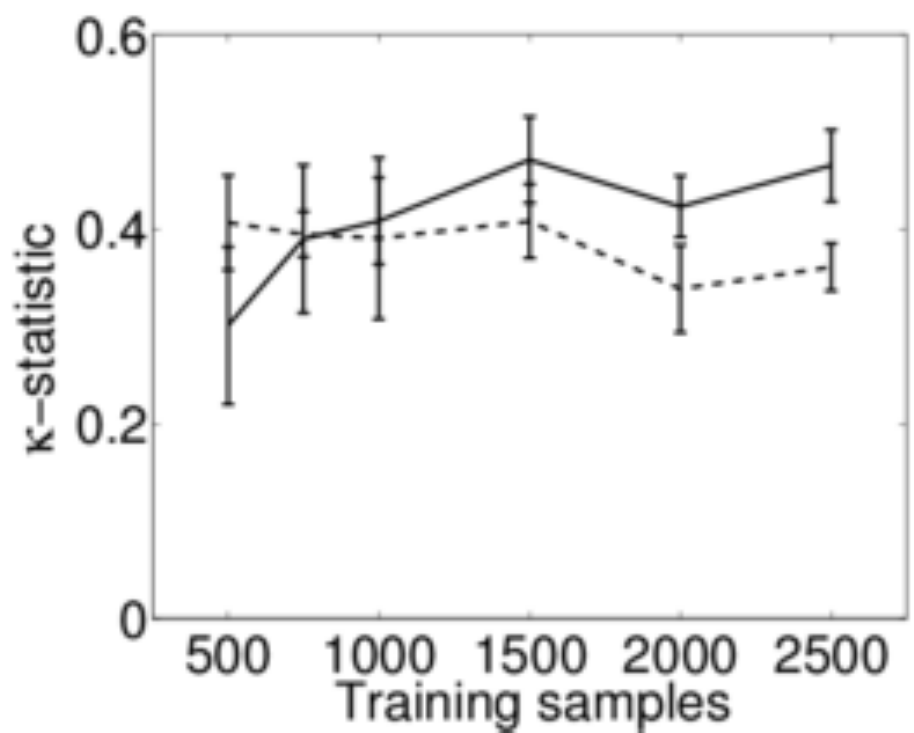}\\
\includegraphics[width=0.45\columnwidth]{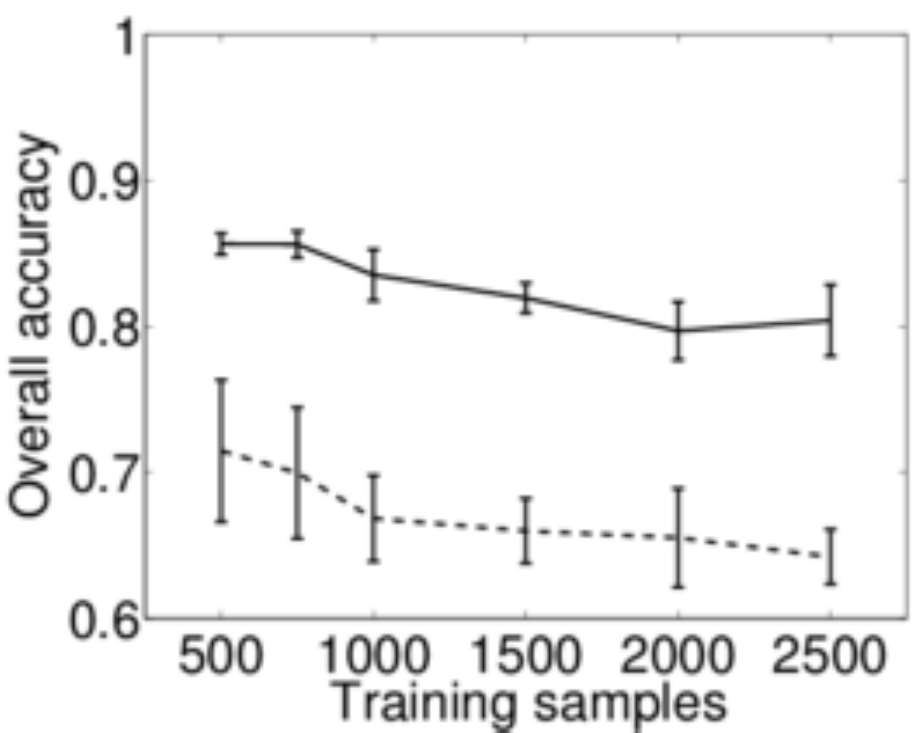}  &   \includegraphics[width=0.45\columnwidth]{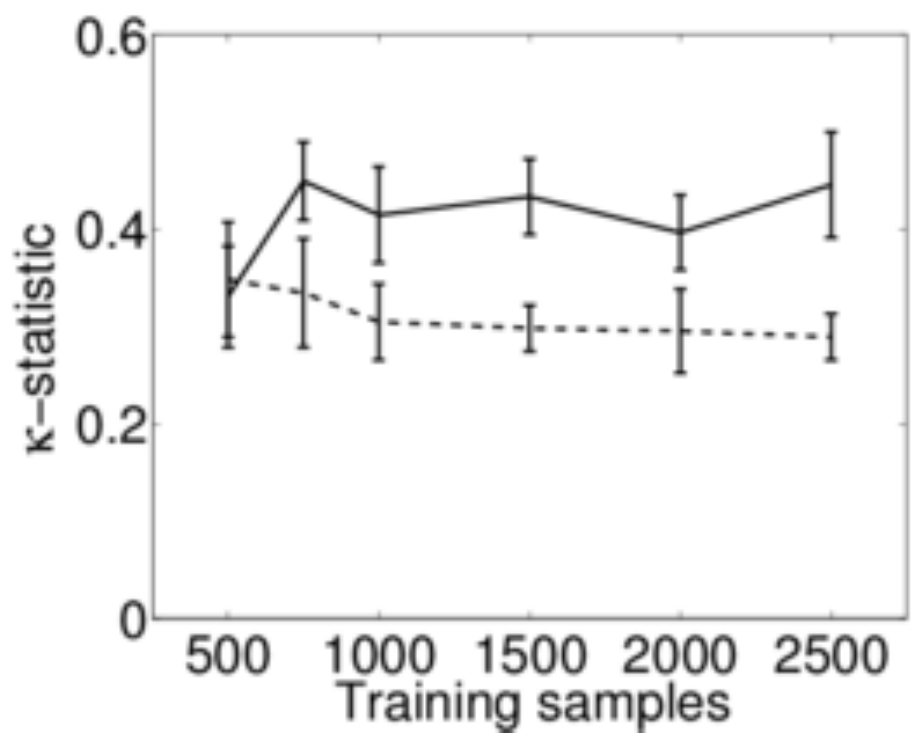} \\
\includegraphics[width=0.45\columnwidth]{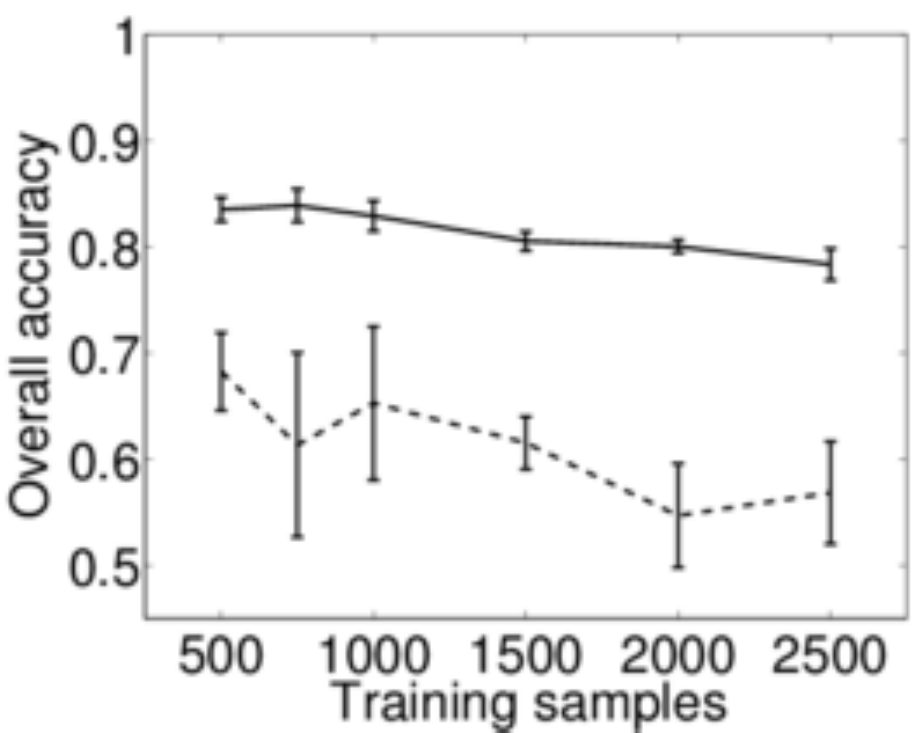} &  \includegraphics[width=0.45\columnwidth]{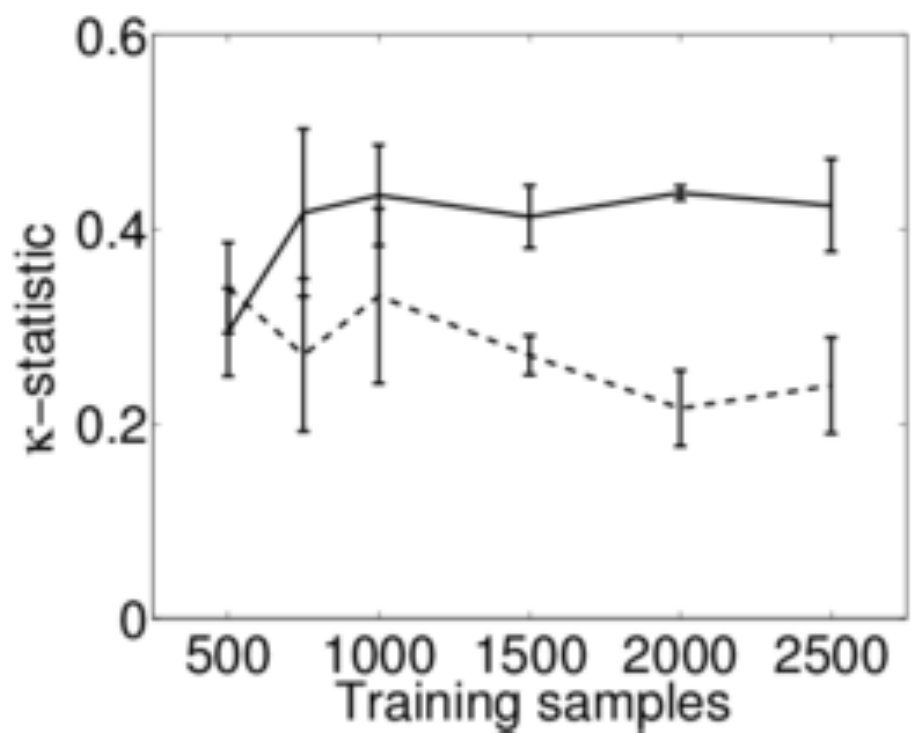}
\end{tabular}
\end{center}
\caption{Overall accuracy (left) and kappa statistic, $\kappa$ (right) for RBIG (solid line) and SVDD (dashed line) in different scenes: Naples 1995 (top), Naples 1999 (center) and Rome 1995 (bottom).}
\label{fig:kappas}
\end{figure}

Figure \ref{fig:kappas} shows the estimated $\kappa$ statistic and the overall accuracy (OA) in the test set achieved by SVDD and RBIG in the three images. The $\kappa$ scores are relatively small because samples were taken from a large spatial area thus giving rise to a challenging problem due to the variance of the spectral signatures. Results show that SVDD behavior is similar to the proposed method for small size training sets. This is because more target samples are needed by the RBIG for an accurate PDF estimation. However, for moderate and large training sets the proposed method substantially outperforms SVDD. Note that training size requirements of RBIG are not too demanding: using $750$ samples in a $10$-dimensional problem is enough for RBIG to outperform SVDD when very little is known about the non-target class.

\begin{figure*}[t!]
\scriptsize
\begin{center}
 \begin{tabular}{ccc}
{\bf GT} & {\bf SVDD (0.67, 0.32)} & {\bf RBIG (0.79, 0.42)} \\
\includegraphics[width=5.5cm]{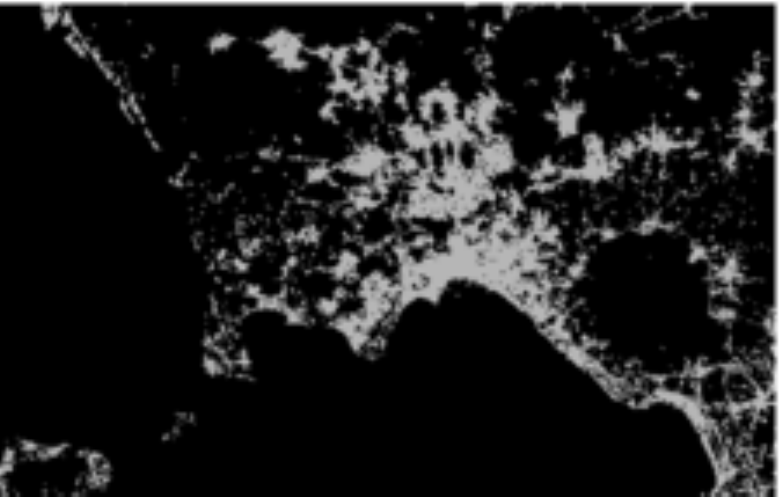}  &
\includegraphics[width=5.5cm]{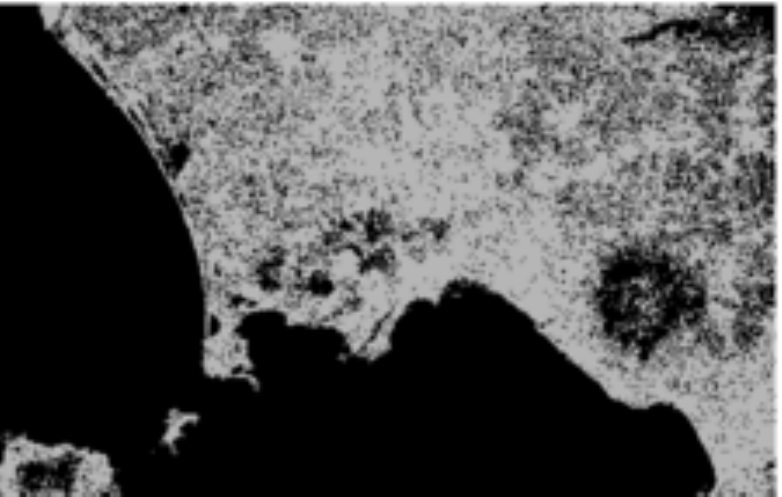}  &
\includegraphics[width=5.5cm]{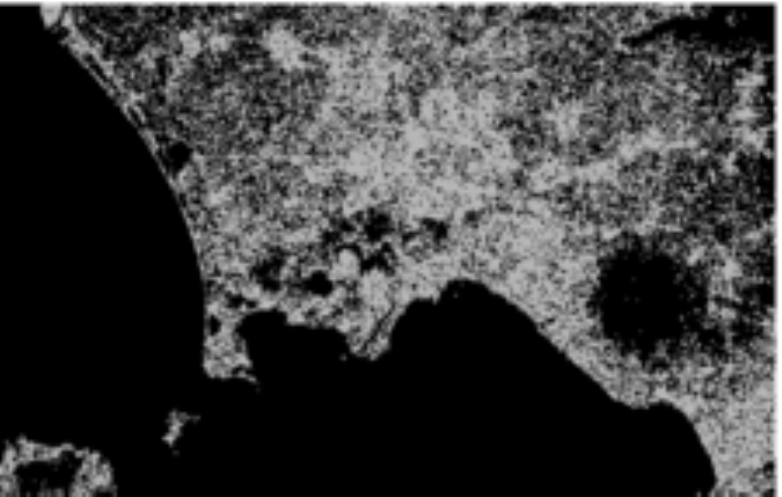}  \\
 \end{tabular}
\end{center}
 \caption{Ground-truth (GT) and classification maps obtained with SVDD and RBIG for the Naples 1995 scene. The white points represent urban area and the black points represent non-urban area. The corresponding overall accuracy and $\kappa$-statistic are given in parenthesis.}
\label{fig:maps}
\end{figure*}

Figure \ref{fig:maps} shows the classification maps for the representative Naples95 scene for SVDD and RBIG. Note that
RBIG better rejects the `non-urban' areas (in black). This may be because SVDD training with few non-target data gives rise to a too broad boundary. As a result, too many pixels are identified as belonging to the target class (in white). Another relevant observation is the noise in neighboring pixels, which may come from the fact that no spatial information was used. This problem could be easily alleviated by imposing some post-classification smoothness constraint or by incorporating spatial texture features.

\subsection{Image Denoising}

Image denoising tackles the problem of estimating the underlying image, $\mathbf{x}$, from a
noisy observation, $\mathbf{x}_n$, assuming an additive degradation model:
$\mathbf{x}_n = \mathbf{x}+\mathbf{n}$. Many methods have exploited the Bayesian framework to this end \cite{Portilla03,Donoho95,simoncelli99b,Figueiredo01}:
\begin{equation}
     \hat{\mathbf{x}} = \underset{\mathbf{x}^*}{\operatorname{argmin}} \bigg\{\int{{\mathcal L}(\mathbf{x},\mathbf{x^*})p(\mathbf{x}|\mathbf{x}_n)d\mathbf{x}}\bigg\},
\label{eq:den_Bayes_risk}
\end{equation}
where $\mathbf{x^*}$ is the candidate image, ${\mathcal L}(\mathbf{x},\mathbf{x^*})$ is the cost function, and $p(\mathbf{x}|\mathbf{x}_n)$ is the posterior probability of the original sample $\mathbf{x}$ given the noisy sample $\mathbf{x}_n$. This last term plays an important role since it can be decomposed (using the Bayes rule) as
\begin{equation}
     p(\mathbf{x}|\mathbf{x}_n) = Z^{-1} p(\mathbf{x}_n|\mathbf{x})p(\mathbf{x}),
\label{eq:den_Bayes}
\end{equation}
where $Z^{-1}$ is a normalization term, $p(\mathbf{x}_n|\mathbf{x})$ is the noise model (probability of the noisy sample given the original one), and $p(\mathbf{x})$ is the prior (marginal) sample model.

Note that, in this framework, the inclusion of a feasible image model, $p(\mathbf{x})$, is critical in order to obtain a good estimation of the original image. Images are multidimensional signals whose PDF $p(\mathbf{x})$ is hard to estimate with traditional methods.
The conventional approach consists of using parametric models to be plugged into Eq. \eqref{eq:den_Bayes} in such a way that the problem can be solved analytically.
However, mathematical convenience leads to the use of too rigid image models.
Here we use RBIG in order to estimate the probability model of natural images $p({\bf x})$.

In this illustrative example, we use the $L_2$-norm as cost function, ${\mathcal L}(\mathbf{x},\mathbf{x^*}) = ||\mathbf{x}-\mathbf{x^*}||_2$, and an additive Gaussian noise model, $p({\bf x}_n|{\bf x}) = {\mathcal N}(\mathbf{0},\sigma_n^2{\bf I})$. We estimated $p(\mathbf{x})$ using $100$ achromatic images of size $256\times 256$ extracted from the McGill Calibrated Colour Image Database\footnote{http://tabby.vision.mcgill.ca/}.
To do this, images were transformed using orthonormal QMF wavelet domain with four frequency scales
\cite{Simoncelli90}, an then each subband was converted to patches in order to obtain different PDF models for each subband according to well-known properties of natural images in wavelet domains \cite{Liu01,Laparra10a}. In order to evaluate Eq. \eqref{eq:den_Bayes_risk}, we sampled the posterior PDF at $8,000$ points from the neighborhood of each wavelet coefficient by generating samples with the PDF of the noise model ($p(\mathbf{x}_n|\mathbf{x})$), and evaluated the probability for each sample with the PDF obtained in the training step $p(\mathbf{x})$.
The estimated coefficient $\hat{\mathbf{x}}$ is obtained as the expected value over the $8000$ samples of the posterior PDF. Obtaining the expected value is equivalent to using the $L_2$ norm \cite{Bernardo}. Note that the classical hard-thresholding (HT) and soft-thresholding (ST) results \cite{Donoho95} are a useful
reference since they can be interpreted as solutions to the same problem with a marginal Laplacian image model
and $L_1$ and $L_2$ norms respectively \cite{simoncelli99b}.

\begin{figure*}[t!]
\small
\begin{center}
\setlength{\tabcolsep}{1pt}
 \begin{tabular}{ccc}
& \hspace{-5cm}{\bf Original} & \hspace{-5cm}{\bf Noisy {(10.0, 0.76)}} \\
& \hspace{-5cm}\includegraphics[width=6cm]{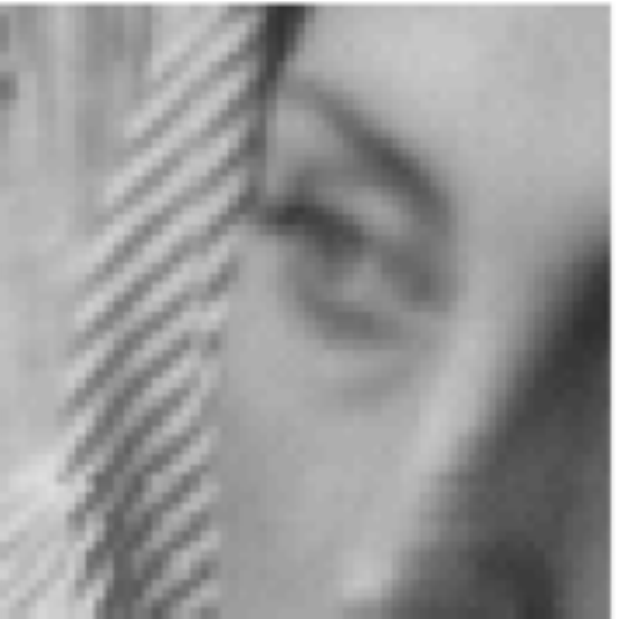}  &
\hspace{-5cm}\includegraphics[width=6cm]{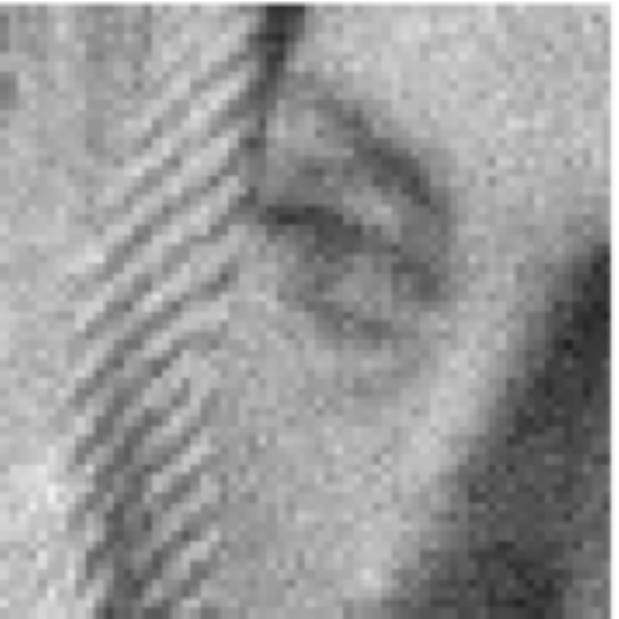} \\
{\bf HT (8.05, 0.86)} & {\bf ST (6.89, 0.88)} & {\bf RBIG (6.48, 0.90)} \\
 \includegraphics[width=6cm]{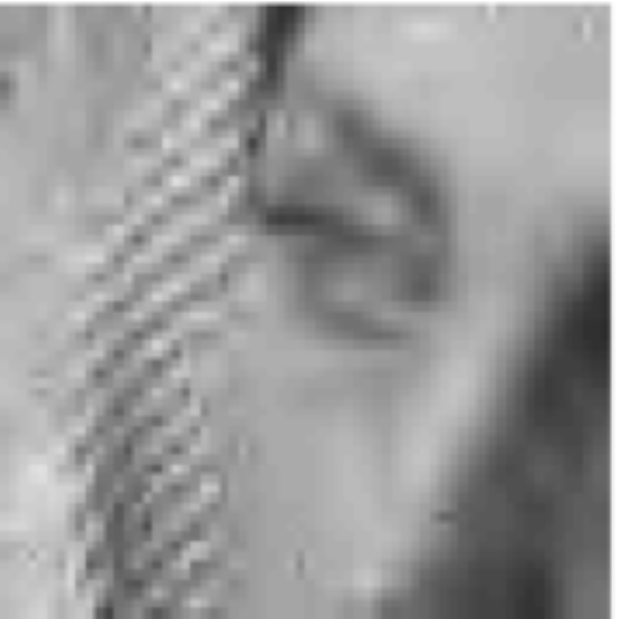}  &
 \includegraphics[width=6cm]{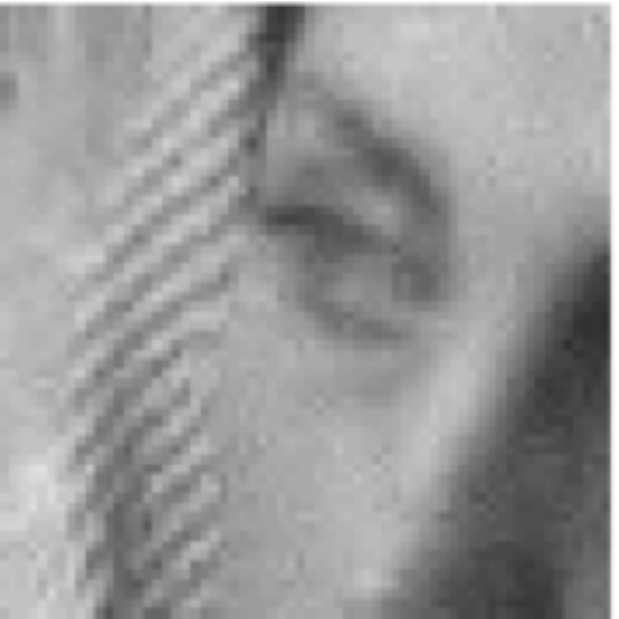}  &
 \includegraphics[width=6cm]{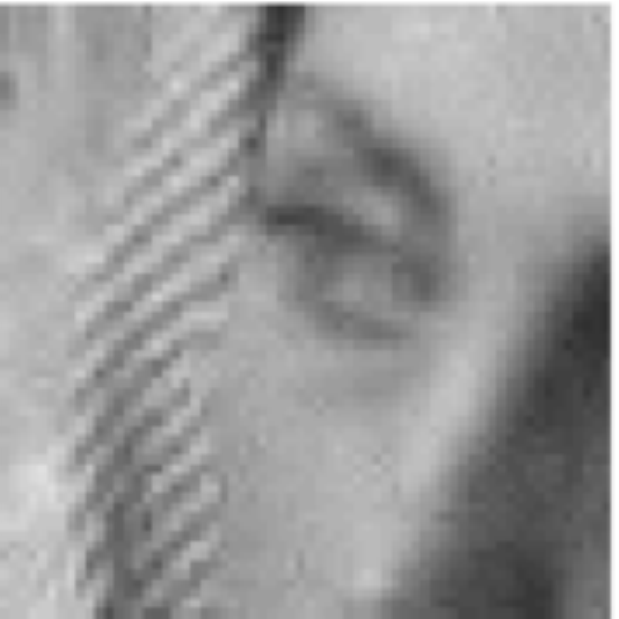}
 \end{tabular}
 \end{center}
 \caption{Original, noisy (noise variance $\sigma_n^2=100$) and restored `Barbara' images. The
root-mean-square-error (RMSE) and the perceptually meaningful Structural Similarity Measure (SSIM) \cite{Wang04}
are given in parentheses.}
\label{den_ori}
\end{figure*}

Figure \ref{den_ori} shows the denoising results for the `Barbara' image corrupted with Gaussian noise of
$\sigma^2_n = 100$ using marginal models (HT and ST), and using a RBIG as the PDF estimator.
Accuracy of the results is measured in Euclidean terms (RMSE), and using a perceptually meaningful image
quality metric such as the Structural Similarity Index (SSIM) \cite{Wang04}.
Note that RBIG method obtains better results (numerically and visually) than the
classical methods due to the more accurate PDF estimation.

\section{Conclusions}
\label{Conclusions}

In this work, we proposed an alternative solution to the PDF estimation problem by using a
family of Rotation-based Iterative Gaussianization (RBIG) transforms.
The proposed procedure looks for differentiable transforms to a Gaussian so
that the unknown PDF can be computed at any point of the original domain using the
Jacobian of the transform.

The RBIG transform consists of the iterative application of univariate
marginal Gaussianization followed by a rotation.
We show that a wide class of orthonormal transforms (including trivial random rotations)
is well suited to Gaussianization purposes.
The freedom to choose the most convenient rotation is the difference with formally
similar techniques, such as Projection Pursuit, focused on looking for interesting
projections (which is an intrinsically more difficult problem). In this way, here we
propose to shift the focus from ICA to a wider class of rotations
since interesting projections as found by ICA are not critical to solve the PDF estimation problem in
the original domain.
The suitability of multiple rotations to solve the PDF estimation problem may help to revive
the interest of classical iterative Gaussianization in practical applications.
As an illustration, we showed promising results in a number of multidimensional
problems such as image synthesis, classification, denoising, and multi-information
estimation.

Particular issues in each of the possible applications, such as stablishing a
convenient family of rotations for a good Jacobian or convenient criteria to
ensure the generalization ability, are a matter for future research.

\section{Acknowledgments}
\label{acks}

The authors would like to thank Matthias Bethge for his constructive criticism to the work, and Eero Simoncelli for the estimulating discussion on `meaningful-vs-meaningless transforms'.

\bibliographystyle{IEEEbib}

\end{document}